\documentclass[11pt,a4paper]{article}

\usepackage{authblk}
\usepackage{natbib}
\usepackage[utf8]{inputenc} 
\usepackage[T1]{fontenc}    
\usepackage{textcomp}
\usepackage{hyperref}       
\usepackage{url}            
\usepackage{booktabs}       
\usepackage{amsfonts}       
\usepackage{nicefrac}       
\usepackage{microtype}      
\usepackage{amsthm}
\usepackage{amsmath,amsfonts,amssymb}
\usepackage{algorithm}
\usepackage{algorithmic}
\usepackage{mathtools}
\usepackage{graphicx}
\usepackage{caption}
\usepackage{subcaption}
\usepackage{subfiles}
\newcommand{\norm}[1]{\left\lVert#1\right\rVert}
\newcommand\inner[2]{\langle #1, #2 \rangle}
\newtheorem{theorem}{Theorem}
\newtheorem{corollary}{Corollary}
\DeclareMathOperator{\vect}{vec}
\DeclareMathOperator{\rank}{rank}
\DeclareMathOperator*{\argmin}{arg\,min}
\newcommand{\var}{\mathrm{var}}
\newcommand{\col}{\mathrm{col}}

\newtheorem{lemma}{Lemma}

\title{\bf Projection-Free Algorithms in Statistical Estimation}

\author[1]{Yan Li}
\author[2]{Chao Qu}
\author[1]{Huan Xu}
\affil[1]{H. Milton Stewart School of Industrial and Systems Engineering, Georgia Institute of Technology}
\affil[2]{Department of Electrical Engineering, Technion }

\begin{document}

\maketitle

\begin{abstract}
  Frank-Wolfe algorithm (FW) and its variants have gained a surge of interests in machine learning community due to its projection-free property.  Recently people have
  reduced the gradient evaluation complexity of FW algorithm  to  $\log(\frac{1}{\epsilon})$ for the smooth and strongly convex objective.
  This complexity result is especially significant in learning problem, as the overwhelming data size makes a single evluation of gradient computational expensive.
  However, in high-dimensional statistical estimation problems, the objective is typically not strongly convex, and sometimes  even non-convex.  In this paper, we extend the
  state-of-the-art FW type algorithms for the large-scale, high-dimensional estimation problem. We show that as long as the objective satisfies {\em restricted strong convexity}, and we are not optimizing over statistical limit of the model, the
  $\log(\frac{1}{\epsilon})$
  gradient evaluation  complexity could still be attained.
\end{abstract}

\section{Introduction}
High-dimensional statistics has achieved remarkable success in the last decade, including results on consistency and convergence
rates for various estimators under non-asymptotic high-dimensional scaling, especially when the problem dimension $p$ is larger
than the number of data $n$.
It is now well known that while this setup appears ill-posed, the estimation or recovery is indeed possible by exploiting
the underlying structure of the parameter space. Notable examples include sparse vectors, low-rank matrices, and structured regression functions, among others.
To impose structural assumption, a popular approach is to do the structural risk minimization, which could be generally formulated as the following optimization problem:
\begin{align}\label{prob_formulation}
\displaystyle
\min_{\varphi(\theta) \leqslant \rho} f(\theta)=\frac{1}{n} \sum_{i=1}^n f_i(\theta)
\end{align}
where $\varphi(\cdot)$ denotes a regularizer and $\rho$ controls the strength of the regularization. We  denote $\Omega = \{\theta: \varphi(\theta) \leqslant \rho\}$ as the constraint set of our problem.
First-order optimization methods for the structural risk minimization problem have been studied extensively in recent years with fruitful developments. (Projected) gradient descent (PGD) algorithm \citep{composite_minimize_nesterov}
and its variants are arguably the most popular algorithm used in practise. It is well-known that projected gradient descent algorithm has $\mathcal{O}(\frac{1}{\epsilon})$
iteration complexity for smooth convex objective, and $\mathcal{O}\left (\log(\frac{1}{\epsilon})\right )$ for smooth and strongly convex objective.

When solving (\ref{prob_formulation}) with batch version of the first-order algorithm, a full pass on the entire dataset to evaluate the gradient could be computational expensive.
In comparison, the stochastic gradient descent (SGD) algorithm computes a noisy gradient over one or a minibatch of samples, which has light computational overhead. However it is known that SGD converges with significantly worse speed \citep{bottou_sgd}.
 Stochastic variance-reduced algorithms \citep{saga_defazio,svrg_zhang,sdca_shalev} combine the merits of both worlds, which has light overhead for computing gradient as SGD, and fast convergence as batch algorithm.

We should note that there are in fact two important operations that consume the major computation resources for PGD algorithm, one being gradient evaluation, and the other one being projection. Although variance-reduced PGD variants reduce complexity of the former, it \emph{does not} improve the latter.
Though commonly assumed to be easy, there are indeed numerous applications where the
projection is computational expensive (projection onto the trace norm ball, base
polytopes \citep{Fujishige_asubmodular}) or even
intractable \citep{collins_intractable}.   Frank-Wolfe (FW) algorithm \citep{vanilla_FW} arises as a
natural alternative in these scenarios. Unlike projection-based algorithm, FW
assumes a linear oracle (LO) that solves a linear optimization problem over the
constraint set, which could be significantly easier than the projection.
However the original Frank-Wolfe algorithm suffers from slow convergence, even when the objective is strongly convex.
It has been shown in \citep{large_scale_lo_lan} that the $\mathcal{O}(\frac{1}{\epsilon})$ LO complexity is not improvable for LO-based algorithm for general smooth and convex problem.
Improving the convergence of FW  is only possible under stronger assumptions such as when the constraint set is polyhedral \citep{linear_fw_julien,stongly_sets_garber}.

The slow sublinear convergence of FW algorithm could be a serious trouble: when applying FW to problem (\ref{prob_formulation}), the algorithm needs an expensive gradient evaluation for every iteration, and the number of iterations needed is also large.
To address this problem, \citet{cgs_lan} show that the gradient evaluation complexity could be improved to $\mathcal{O}(\frac{1}{\sqrt{\epsilon}})$ for the smooth and convex objective, and
$\mathcal{O}(\log(\frac{1}{\epsilon}))$ for the smooth and strongly convex objective.  For the finite sum problem in (\ref{prob_formulation}), similar result  has also been shown  in  \citep{storc_hazan} with variance-reduction technique.
In summary, for both smooth convex problem and smooth strongly convex problem, gradient evaluation complexity of PGD type algorithms and FW type algorithms are comparable, in the sense that they share the same dependency on $\epsilon$ in order to obtain an $\epsilon$-optimal solution.


\textbf{Related Work:} The high-dimensional nature in statistical estimation problem is particularly suited for FW algorithm, as the projection to the constraint set could be difficult. Consider when the constraint set is a  nuclear norm ball, then projection requires performing full SVD, while linear oracle only requires computing the leading singular vectors.
One might assume the lack of strong convexity in the objective would imply sublinear convergence of the PGD algorithm. This is indeed not true.
For the batch PGD algorithm,  \citet{fast_global} show the linear convergence of the algorithm until reaching statistical limit of the model. Their quick convergence result is based on the assumption that objective function satisfies the restricted strong convexity (RSC).
 The batch PGD requires $\mathcal{O}(n \frac{L}{\hat{\sigma}} \log(\frac{1}{\epsilon}))$ gradient evaluations and $\mathcal{O}( \frac{L}{\hat{\sigma}} \log(\frac{1}{\epsilon}))$
projections, where $\hat{\sigma}$ relates to the RSC of objective function.
 \citet{svrg_qu,sdca_qu,saga_qu} later extend the linear convergence result to stochastic variance-reduced PGD algorithms, including  SDCA, SVRG, SAGA. Their results  improve the gradient evaluation complexity to $\mathcal{O}((n +\frac{L}{\hat{\sigma}}) \log(\frac{1}{\epsilon}))$, while the number of
projections remains the same.

\textbf{Contributions:} In contrast to the fruitful study on PGD type algorithm in high dimensional estimation problem (\ref{prob_formulation}), to the best of our knowledge there are still no results on how the FW type algorithm would perform on this problem. Our result in this paper shows that like the PGD type algorithm, the efficiency of FW type algorithm (i.e. CGS and STORC) could be extended much more generally to the setting where objective only satisfies the restricted strong convexity.
 We briefly summarize our (informal) main results and defer the detailed discussions to our formal theorems and corollaries.
\begin{itemize}
  \item Assume $f(\theta)$ is restricted strongly convex, Conditional Gradient Sliding (CGS)
  algorithm attains $\mathcal{O} ( n (\frac{L}{\hat{\sigma}})^{{\frac{1}{2}}} \log(\frac{1}{\epsilon}))$ gradient evaluation complexity up to the statistical limit of the model. Our result matches the known result $\mathcal{O} ( n (\frac{L}{\sigma})^{{\frac{1}{2}}} \log(\frac{1}{\epsilon}))$ which requires strong convexity assumption.
  \item Assume $f(\theta)$ is restricted strongly convex and each $f_i(\theta)$ is convex, stochastic variance-reduced CGS (STORC) algorithm attains
  $\mathcal{O} ( (n + (\frac{L}{\hat{\sigma}})^2) \log(\frac{1}{\epsilon}))$ gradient evaluation complexity up to the statistical limit of the model.
  Our result also matches the known result $\mathcal{O} ( (n + (\frac{L}{\sigma})^2) \log(\frac{1}{\epsilon}))$ which requires strong convexity assumption.
   Similar complexity could also be obtained even if $f_i$ is non-convex.
   \item Both STORC and CGS still maintain the optimal LO complexity $\mathcal{O} (\frac{LD^2}{\epsilon})$ under restricted strong convexity.
  \item The main technical challenge comes from the non-convexity. For instance, to handle the non-convexity in STORC, we exploit the notion of lower smoothness, and generalize the analysis of STORC to a type of non-convex setting.
\end{itemize}

\section{Problem Setup}
We begin with some definitions in the  classic optimization literature.
A function $f(x)$ is  $L$-smooth and $\sigma$-strongly convex with respect to norm $\norm{\cdot}$ if for all $x,y \in dom(f)$:
\begin{align} \frac{\sigma}{2} \norm{x-y}^2  \leqslant f(y)-f(x)- \inner{\nabla f(x)}{ y-x } \leqslant \frac{L}{2} \norm{x-y}^2
\end{align}
we additionally say $f(x)$ is $l$-lower smooth if:
\begin{align}
  -\frac{l}{2} \norm{y-x}^2 \leqslant f(y) -f(x) -\inner{\nabla f(x)}{y-x}.
\end{align}

\textbf{Restricted Strong Convexity.}
The restricted strong convexity was initially proposed in \citep{unified_framework_m-estimator} to establish
statistical consistency of the regularized M-estimator, and later was exploited to
establish linear convergence of the gradient descent algorithm up to the statistical limit of the model \citep{fast_global}. We say $f(x)$ satisfies
restricted strong convexity with respect to $\varphi(x)$ and norm $\norm{\cdot}$ with parameter $(\sigma, \tau_{\sigma})$  if:
\begin{align}\label{rsc_def}
  f(y)-f(x)- & \langle \nabla f(x), y-x \rangle \geqslant
    \frac{\sigma}{2}\norm{y-x}^2 - \tau_{\sigma} \varphi^2(y-x).
\end{align}

\textbf{Decomposable regularizer.}
Intuitively, to ensure restricted strong convexity to be close to strong convexity, one needs to ensure $\varphi(y-x)$ in definition (\ref{rsc_def}) to be close to $\norm{y-x}$. This is an essential argument for establishing consistency of m-estimator and fast convergence of PGD algorithm.
The decomposable regularizer is a sufficient condition to establish such result.
Given a subspace pair $ M \subseteq \overline{M}$ which belongs to a Hilbert space $\mathbf{X}$, define
$
  \overline{M}^{\perp} = \{v \in \mathbf{X} : \inner{v}{u} = 0, \, \forall u \in \overline{M} \}
$ to be the orthogonal complement of $ \overline{M}$.
We say that the regularizer  $\varphi(x)$ is
decomposable with respect to $(M,\overline{M}^{\perp} )$ if:
\begin{align}
  \varphi(x+y)=\varphi(x)+\varphi(y) \quad \forall x\in M, \, y \in \overline{M}^{\perp}
\end{align}
We further define the subspace compatibility as
$
  \phi(M)=\underset{x \in M\setminus \{0\}}{\sup} \frac{\varphi(x)}{||x||}.
$.

\textbf{Assumptions.}
We will make the following assumptions throughout this paper. We assume each $f_i$ is $L$-smooth, hence $f$ is also $L$-smooth. If $f_i$ is non-convex, we will assume it is $l$-lower smooth. Finally, we assume that $f$ satisfies
restricted strong convexity with respect to $\varphi$ with parameter $(\sigma, \tau_{\sigma})$.

\subsection{Example: Matrix Regression}
In this subsection we put previously defined notions into context and
consider the matrix regression problem: let $\Theta^{\star}\in \mathbb{R}^{d\times d}$ be an unknown low-rank matrix to be
estimated and we have observations from the model:
$
   y_i = \inner{ X_i} {\Theta^{\star}}  + \epsilon_i,
$
where $\epsilon_i$ is i.i.d sampled from $N(0, \nu^2)$. We denote $\inner{X}{Y } = \mathrm{tr}(X^T Y)$ as the  Frobenius inner product
 and  $\norm{X}_F =\sqrt{ \mathrm{tr}(X^T X)}$ as the Frobenius norm of matrix $X$.

\textbf{Decomposibility of nuclear norm.}
Define $\norm{X}_* = \mathrm{tr}(\sqrt{X^{T} X})$ to be the nuclear norm of matrix $X$, then it is well known that $\norm{X + Y}_* = \norm{X}_* + \norm{Y}_*$, if $X$ and $Y$ have orthogonal row space and column space.
 Let $\Theta^{\star} = UDV^T$ be the singular value
decomposition of $\Theta^{\star}$. Define $V^r$ as the submatrix of $V$ with the first $r$ columns and similarly define $U^r$. Let $\col(X)$ denote the column space of a matrix $X$, we define the subspace pair:
\begin{align}
 \mathcal{M}(U^r,V^r) & = \{\Theta:  \col(\Theta^T) \subseteq \col(V^r), \col(\Theta) \subseteq \col(U^r) \} \nonumber \\
 \overline{\mathcal{M}}^{\perp}(U^r,V^r) & = \{ \Theta :\col(\Theta^T) \subseteq \col(V^r)^{\perp},
 \col(\Theta) \subseteq \col(U^r)^{\perp} \} \nonumber
\end{align}
Then any $\Theta \in \mathcal{M}(U^r,V^r)$ and $\Gamma \in \overline{\mathcal{M}}^{\perp}(U^r,V^r)$ have orthogonal row and column subspaces, hence we have the decomposibility of nuclear norm with respect to $(\mathcal{M}, \overline{\mathcal{M}}^{\perp} )$.

We will use $\vect(M)$ to denote the vectorization of  matrix $M$.
We define the loss function for matrix regression in (\ref{loss_mat}), the $\hat{\Gamma}$ and $\hat{\gamma}$ will be specified seperately depending on  whether the sensing matrix $X_i$ is observed with noise.
\begin{align}\label{loss_mat}
  L_n(\Theta) = \frac{1}{2} \vect(\Theta)^T \hat{\Gamma} \vect(\Theta) - \inner{\hat{\gamma}}{\vect(\Theta)}
\end{align}

\textbf{Convex loss:} If sensing matrix $X_i$ is observed without noise, we set $\hat{\Gamma} = \frac{1}{n} X^T X$ and $\hat{\gamma} = \frac{1}{n} X^T y$, where $\vect(X_i)$ is the $i$-th row of the matrix $X \in \mathbb{R}^{n \times d^2}$.
It is easy to verify that,
\begin{align}\label{mat_convex_loss}
  L_n(\Theta) =\frac{1}{2n} \sum_{i=1}^n \left(y_i - \inner{X_i}{ \Theta} \right)^2
\end{align}

\textbf{Non-convex loss:} If the sensing matrix $X_i$ is observed with additive noise, i.e. we observe $Z_i = X_i +W_i$ where $W_i$ is independent with $X_i$ and $\vect(W_i) \sim N(0,\Sigma_w)$. We set $\hat{\Gamma} = \frac{1}{n} Z^TZ -\Sigma_w$ and $\hat{\gamma} = \frac{1}{n} Z^T y$, where $\vect(Z_i)$ is the $i$-th row of the matrix $Z \in \mathbb{R}^{n \times d^2}$. We could rewrite (\ref{loss_mat}) as:
\begin{align}\label{mat_nonconvex_loss}
  L_n(\Theta) = \frac{1}{2n} \sum_{i=1}^n\left \{ \left(y_i - \inner{Z_i}{\Theta} \right)^2 - \vect(\Theta)^T \Sigma_w \vect(\Theta) \right \}
\end{align}
Note that $\rank(Z^TZ) \leqslant n$ and we subtract a positive definite matrix $\Sigma_w \in \mathbb{R}^{d^2 \times d^2}$ from it
, hence $\hat{\Gamma}$ can not be positive semidefinite when $n < d^2$ (the typical setup for matrix regression),
which results in the non-convexity of $L_n(\Theta)$.

\textbf{Structural risk minimization:} To impose low-rankness of the recovered matrix, a popular approach is to minimize the objective over a  nuclear norm ball with suitable radius
 \citep{matrix_completion_koltchinskii}. The matrix regression problem is defined by:
\begin{align}\label{matrix_regression}
\hat{\Theta}  = \argmin_{\norm{\Theta}_* \leqslant \rho} L_n(\Theta)
\end{align}
Lamma \ref{rsc_lemma} establishes restricted strong convexity of the loss function in the matrix regression problem, which was essentially proved in \citep{fast_global}.
\begin{lemma}\label{rsc_lemma}
  Suppose sensing matrix $X_i$ is i.i.d. sampled from $N(0, \Sigma_x)$
 , define $\xi(\Sigma_x) = \sup_{\norm{u}=1, \norm{v}=1} \var( u^T X_i v )$, we have for loss function (\ref{mat_convex_loss}):
\begin{align*}
  L_n(V) & - L_n (U) - \inner{\nabla L_n(U)}{V-U} \geqslant
   \frac{\lambda_{\min}(\Sigma_x)}{2} \norm{V-U}_F^2  - \frac{c \xi(\Sigma_x) d}{n} \norm{V-U}_{*}^2
\end{align*}
Suppose $\vect(W_i)$ is i.i.d. sampled from $N(0, \Sigma_w)$ and independent with $X_i$.
Assume that $\lambda_{max}(\Sigma_w) \leqslant \frac{1}{4} \lambda_{min}(\Sigma_x)$, we have for loss function (\ref{mat_nonconvex_loss}):
\begin{align*}
  L_n(V) & - L_n (U) - \inner{\nabla L_n(U)}{V-U} \geqslant
   \frac{\lambda_{\min}(\Sigma_x)}{4} \norm{V-U}_F^2  - \frac{c \xi(\Sigma_x) d}{n} \norm{V-U}_{*}^2
\end{align*}
with probability at least $1- c_1 \exp(-c_2 n)$, for some absolute constants $c, c_1, c_2$.
\end{lemma}

\section{Theoretical Results}
In this section we present the complexity results for CGS and STORC, under the restricted strongly convex assumption.
A few more definition is needed for presenting our formal results.

\textbf{Definition.}
Let $f$ be a function satisfying restricted strong convexity with respect to $\varphi$ with parameter $(\sigma, \tau_{\sigma})$. Suppose $\varphi$ is decomposable
with respect to subspace pair $(\mathcal{M}, \overline{\mathcal{M}}^{\perp})$ and let $\Pi_{\mathcal{M}^{\perp}}(\cdot)$ denotes projection onto subspace $\mathcal{M}^{\perp} $. Let $\theta^{\star}$ be the unknown target parameter, $\hat{\theta}$ be the optimal solution
for (\ref{prob_formulation}). We define the following quantity:
\begin{itemize}
  \item Optimization error: $\Delta^{\star}=\hat{\theta}-\theta^{\star}$;
  \item Effective strong convexity parameter: $\hat{\sigma}=(\sigma - 16\tau_{\sigma} \phi(\overline{\mathcal{M}})^2) $;
  \item Statistical error: $\epsilon_{stat}^2 = \tau_{\sigma}(\varphi(\Pi_{\mathcal{M}^{\perp}}(\theta^{\star}))+2\varphi(\Delta^{\star})+\phi(\overline{\mathcal{M}})\norm{\Delta^{\star}})^2/\hat{\sigma}$.
\end{itemize}

\textbf{Parameter Specification.}
  In the CGS and STORC algorithm, we set:
  \begin{align}\label{pr_1}
    N_t = 8 \sqrt{\frac{L}{\hat{\sigma}}}, \gamma_k = \frac{2}{k+1}, \beta_k = \frac{3L}{k}, \eta_{t,k} = \frac{8L\delta_0 2^{-t}}{\hat{\sigma} N_t k };
  \end{align}
 for convex STORC we set:
 \begin{align}\label{pr_2}
 m_{t,k}=  \frac{5200 N_t L}{\hat{\sigma}};
\end{align}
and for non-convex STORC we set:
\begin{align}\label{pr_3}
\tilde{L} = (L+l)(1+\frac{l}{\hat{\sigma}}), \,\, m_{t,k}=  \frac{8000 N_t \tilde{L}}{\hat{\sigma}},
\end{align}

We present the batch algorithm CGS \citep{cgs_lan} in Algorithm \ref{alg:CGS}. CGS is a smart combination of Nesterov's accelerated gradient descent (AGD) \citep{agd_nesterov} and the Frank-Wolfe algorithm, which essentially uses FW algorithm to solve the projection subproblem in AGD.

\begin{algorithm}[]
   \caption{Conditional Gradient Sliding (CGS)}
   \label{alg:CGS}
\begin{algorithmic}
   \STATE {\bfseries Input:} objective function $f(\theta)=\frac{1}{n}\sum_{i=1}^n f_i(\theta)$, parameters $\gamma_k, \beta_k, \eta_{t,k}, N_t$
   \STATE {\bfseries Initialize:} Choose any $\theta_0$ that $\varphi(\theta_0) \leqslant \rho$.
   \FOR{$t=1,2\ldots$}
   \STATE Let $x_0=y_0=\theta_{t-1}$
   \FOR{$k=1$ to $N_t$}
   \STATE $z_k=(1-\gamma_k) y_{k-1} + \gamma_k x_{k-1}$
   \STATE Set $\nabla_k = \nabla f(z_k)$
   \STATE Let $g(x)=\frac{\beta_k}{2}\norm{x-x_{k-1}}^2 + \nabla_k^T x$
   \STATE Solve subproblem $\underset{\varphi(x) \leqslant \rho}{\text{min}} g(x)$ using
   Frank-Wolfe algorithm, output $x_k$ such that $\underset{\varphi(x) \leqslant \rho}{\text{max}} \inner{\nabla g(x_k)}{x_k -x} \leqslant \eta_{t,k} $
   \STATE $y_k=(1-\gamma_k)y_{k-1}+\gamma_k x_k$
   \ENDFOR
   \STATE Let $\theta_t=y_{N_t}$
   \ENDFOR
\end{algorithmic}
\end{algorithm}

We now present our result for Conditional Gradient Sliding. We emphasize that our result holds uniformly for both convex and non-convex setting, with  parameter setting that does not depend on convexity.
\begin{theorem}\label{thrm:CGS}
  Let $f, f_i$ satisfy our assumptions and be restricted strongly convex with respect to $\varphi$ with parameter $(\sigma, \tau_{\sigma})$. Suppose $\varphi$ is decomposable
  with respect to the subspace pair $(\mathcal{M}, \overline{\mathcal{M}}^{\perp})$ and $\varphi(\hat{\theta})=\rho$. Let $\delta_0 $ be an estimate such that $f(\theta_0)-f(\hat{\theta}) \leqslant \delta_0$, $D$ be the diameter of feasible set $\Omega$. If we run CGS with parameter specified in (\ref{pr_1}), then for both the convex and the non-convex case: for any $\epsilon \geqslant \hat{\sigma}\epsilon_{stat}^2$, in order for the  CGS  algorithm to obtain an iterate $\theta_t$ such that $f(\theta_t)-f(\hat{\theta})\leqslant \epsilon$, the number of calls to gradient evaluation and linear oracles are bounded respectively
  by:
  \begin{equation}\label{batch_grad}
    \mathcal{O}\left (n\sqrt{\frac{L}{\hat{\sigma}}} \log_2(\frac{\delta_0}{\epsilon})\right ),
  \end{equation}
  and
  \begin{equation}\label{batch_lo}
    \mathcal{O}\left (\frac{LD^2}{\epsilon}+\sqrt{\frac{L}{\hat{\sigma}}} \log_2(\frac{\delta_0}{\epsilon}) \right ).
  \end{equation}
  Furthermore, take $\epsilon = \hat{\sigma} \epsilon_{stat}^2$, whenever $f(\theta_t)-f(\theta) \leqslant \epsilon$ holds, we would have:
  \begin{equation}
    \norm{\theta_t-\hat{\theta}}^2 \leqslant \epsilon_{stat}^2.
  \end{equation}
\end{theorem}
\textbf{Remarks:}
We note that these bounds are known to be optimal for smooth and strongly convex function \citep{cgs_lan}, and we extend  the applicability of CGS in the following sense:
the bounds also hold for a class of convex but not strongly convex functions and even non-convex functions, provided that they satisfy the restricted strong convexity. While our result
has a mild restriction on the precision up to which the algorithm converges at the  predicted rate, we remark that there would be no additional statistical gains for optimizing over this precision. In fact in many models, $\epsilon_{stat}^2$ is shown to be on the same or lower order of the statistical precision of the model, as to be illustrated in the following corollary.

\begin{corollary}\label{batch_cor}
For both convex and non-convex matrix regression problem (\ref{matrix_regression}). Assume that $\rho \leqslant \varphi(\theta^{\star})$, then under the condition of Lemma \ref{rsc_lemma},  we have  $\epsilon_{stat}^2 = \frac{\xi(\Sigma_x) d r}{\hat{\sigma} n} \norm{\Delta^{\star}}_F^2$,
$L= \sigma_{max}(\Sigma_x)$ and $\hat{\sigma} =  \sigma_{min}(\Sigma_x) - c \frac{\xi(\Sigma_x)dr}{n}$ for an absolute constant $c$.
For sample size satisfying the scaling
$n = 2c\frac{\xi(\Sigma_x ) d r}{\sigma_{min}(\Sigma_x)}$, we have $\hat{\sigma} = \frac{ \sigma_{min}(\Sigma_x)}{2}$.
Let $\epsilon = \frac{\hat{\sigma} \xi(\Sigma_x) d r }{\sigma_{min}(\Sigma_x)n} \norm{\Delta^{\star}}_F^2$,
then with
\begin{align}
  \mathcal{O} \left ( n \sqrt{\frac{\sigma_{max}(\Sigma_x)}{\sigma_{min}(\Sigma_x) }} \log_2(\frac{\delta_0}{\epsilon})  \right)
\end{align}
 gradient evaluations, and
\begin{align}
  \mathcal{O} \left(\frac{\sigma_{max}(\Sigma_x) \rho^2}{\epsilon} + \sqrt{ \frac{\sigma_{max}(\Sigma_x)}{\sigma_{min}(\Sigma_x)  }}  \log_2(\frac{\delta_0}{\epsilon})  \right)
\end{align}
calls to the linear oracle, Algorithm~\ref{alg:CGS} achieves an optimaliy gap $f(\theta_t) - f(\hat{\theta}) \leqslant \epsilon$, and the distance to optimum:
\begin{align}
  \norm{\theta_t - \hat{\theta}}_F^2 \leqslant c_2 \cdot \norm{\Delta^{\star}}_F^2
\end{align}
where $c_2$ is an absolute constant.
\end{corollary}

Next we present the Stochastic Variance Reduced Conditional Gradient Sliding \citep{storc_hazan}. Its only difference from the CGS algorithm is to replace the full gradient $\nabla_k$
with a variance-reduced gradient in SVRG style. We use $t$ to index the outer iteration, and $k$ to index the inner iteration. We compute the full
gradient for each outer iteration, and for each inner iteration we replace $\nabla_k$ with $\tilde{\nabla}_k=\frac{1}{m_t,k} \sum_{j \in J}( \nabla f_j(z_k) -\nabla f_j(y_0) + \nabla f(y_0))$,
where $J$ is the set of $m_{t,k}$ indices sampled uniformly and independently from $\{1,\ldots,n\}$. The details are presented in Algorithm \ref{alg:STORC}.

\begin{algorithm}[tbh]
   \caption{Stochastic Variance Reduced Conditional Gradient Sliding (STORC)}
   \label{alg:STORC}
\begin{algorithmic}
   \STATE {\bfseries Input:} objective function $f(\theta)=\sum_{i=1}^n f_i(\theta)$, parameters $\gamma_k, \beta_k, \eta_{t,k}, N_t$
   \STATE {\bfseries Initialize:} Choose any $\theta_0$ that $\varphi(\theta_0) \leqslant \rho$.
   \FOR{$t=1,2\ldots$}
   \STATE Let $x_0=y_0=\theta_{t-1}$
   \STATE Compute full gradient $\nabla f(y_0)$
   \FOR{$k=1$ to $N_t$}
   \STATE $z_k=(1-\gamma_k) y_{k-1} + \gamma_k x_{k-1}$
   \STATE Sample $J$, compute $\nabla_k = \frac{1}{m_{t,k}} \sum_{j \in J}( \nabla f_j(z_k) -\nabla f_j(y_0) + \nabla f(y_0))$
   \STATE Let $g(x)=\frac{\beta_k}{2}\norm{x-x_{k-1}}^2 + \nabla_k^T x$
   \STATE Solve subproblem $\underset{\varphi(x) \leqslant \rho}{\text{min}} g(x)$ using
   Frank-Wolfe algorithm, output $x_k$ such that $\underset{\varphi(x) \leqslant \rho}{\text{max}} \inner{\nabla g(x_k)}{x_k -x} \leqslant \eta_{t,k} $
   \STATE $y_k=(1-\gamma_k)y_{k-1}+\gamma_k x_k$
   \ENDFOR
   \STATE Let $\theta_t=y_{N_t}$
   \ENDFOR
\end{algorithmic}
\end{algorithm}

\begin{theorem}\label{thrm:STORC}
Under the same conditions as Theorem \ref{thrm:CGS},  if each $f_i$ is convex: run STORC  with parameters specified in (\ref{pr_1}) and (\ref{pr_2}); if $f_i$ is non-convex but is $l$-lower smooth: run STORC  with parameters specified in (\ref{pr_1}) and (\ref{pr_3}).
 Then for any $\epsilon \geqslant \hat{\sigma}\epsilon_{stat}^2$, in order to obtain an iterate $\theta_t$ such that $\mathbb{E} \left [f(\theta_t)-f(\hat{\theta}) \right ]\leqslant \epsilon$, the number of  gradient evaluation is bounded by:
\begin{align}
  \text{Convex:} \quad & \quad \mathcal{O} \left (\left(n+(\frac{L}{\hat{\sigma}} )^2  \right)\log_2(\frac{\delta_0}{\epsilon}) \right ), \\
  \text{Non-convex:} \quad & \quad \mathcal{O} \left (\left(n+\frac{\tilde{L}L}{\hat{\sigma}^2}  \right)\log_2(\frac{\delta_0}{\epsilon}) \right ),
\end{align}
and the number of calls to linear oracle is bounded by:
\begin{equation}
  \mathcal{O} \left( \frac{LD^2}{\epsilon} +\sqrt{\frac{L}{\hat{\sigma}}} \log_2(\frac{\delta_0}{\epsilon}) \right ).
\end{equation}
Furthermore, take $\epsilon = \hat{\sigma} \epsilon_{stat}^2$, whenever $f(\theta_t)-f(\theta) \leqslant \epsilon$ holds, we have:
\begin{equation}\label{staterr_bound}
  \norm{\theta_t-\hat{\theta}} \leqslant \epsilon_{stat}^2,
\end{equation}
\end{theorem}

\textbf{Remarks:} For the convex loss, our result parallels with original STORC, except that we do not need  $f$ to be strongly convex. For the non-convex loss, our result comes from an analysis of STORC applying to a  strongly convex objective (indeed RSC suffices) that is a summation of \textit{non-convex} functions.
Note that similar results have been shown only for the variance-reduced PGD type algorithm \citep{improved_svrg_allen}.
We also note that whenever $l \leqslant \hat{\sigma}$, we have $\tilde{L} \leqslant 4L$, which implies the complexity of the non-convex objective reduces to that of the convex objective. In other words, we pay no penalty for handling non-convexity. Since we can always bound $l$ by $L$, the worst case the gradient evaluation complexity for the non-convex loss is $\mathcal{O} \left( (n + (\frac{L}{\hat{\sigma}})^3) \log_2(\frac{\delta_0}{\epsilon}) \right )  $.

\begin{corollary}\label{stochastic_cor}
  For both convex and non-convex matrix regression problem (\ref{matrix_regression}). Assume $\rho \leqslant \varphi(\theta^{\star})$, then under the conditions of Lemma \ref{rsc_lemma},  we have $\epsilon_{stat}^2 =  \frac{\xi(\Sigma_x) d r}{\hat{\sigma} n} \norm{\Delta^{\star}}_F^2$,
  $L= \sigma_{max}(\Sigma_x)$, $\tilde{L} = \sigma_{max}(\Sigma_x) + \sigma_{min}(\Sigma_x)$ and $\hat{\sigma} =  \sigma_{min}(\Sigma_x) - c \frac{\xi(\Sigma_x)dr}{n}$ for some universal constant $c$.
  Assume the sample size satisfies scaling
  $n =  2c\frac{\xi(\Sigma_x ) d r}{\sigma_{min}(\Sigma_x)} $,  we have $\hat{\sigma} = \frac{ \sigma_{min}(\Sigma_x)}{2}$ .
  Let $\epsilon = \frac{\hat{\sigma} \xi(\Sigma_x) d r }{\sigma_{min}(\Sigma_x)n} \norm{\Delta^{\star}}_F^2$,
  then with
  \begin{align}
    \mathcal{O} \left (  (n + (\frac{\sigma_{max}(\Sigma_x)}{\sigma_{min}(\Sigma_x)})^2 ) \log_2(\frac{\delta_0}{\epsilon})  \right)
  \end{align}
gradient evaluations, and
  \begin{align}
    \mathcal{O} \left(\frac{\sigma_{max}(\Sigma_x) \rho^2}{\epsilon} + \sqrt{ \frac{\sigma_{max}(\Sigma_x)}{\sigma_{min}(\Sigma_x)  }}  \log_2(\frac{\delta_0}{\epsilon})  \right)
  \end{align}
  calls to to the linear oracle, Algorithm~\ref{alg:STORC} achieves an optimality gap $f(\theta_t) - f(\hat{\theta}) \leqslant \epsilon$, and the distance to the optimum satisfies
  \begin{align}
    \norm{\theta_t - \hat{\theta}}_F^2 \leqslant c_2 \cdot \norm{\Delta^{\star}}_F^2
  \end{align}
  where $c_2$ is an absolute constant.
\end{corollary}

\section{Simulation}
\begin{figure}[]
\centering
\begin{subfigure}{.32\textwidth}
\includegraphics[width=\linewidth]{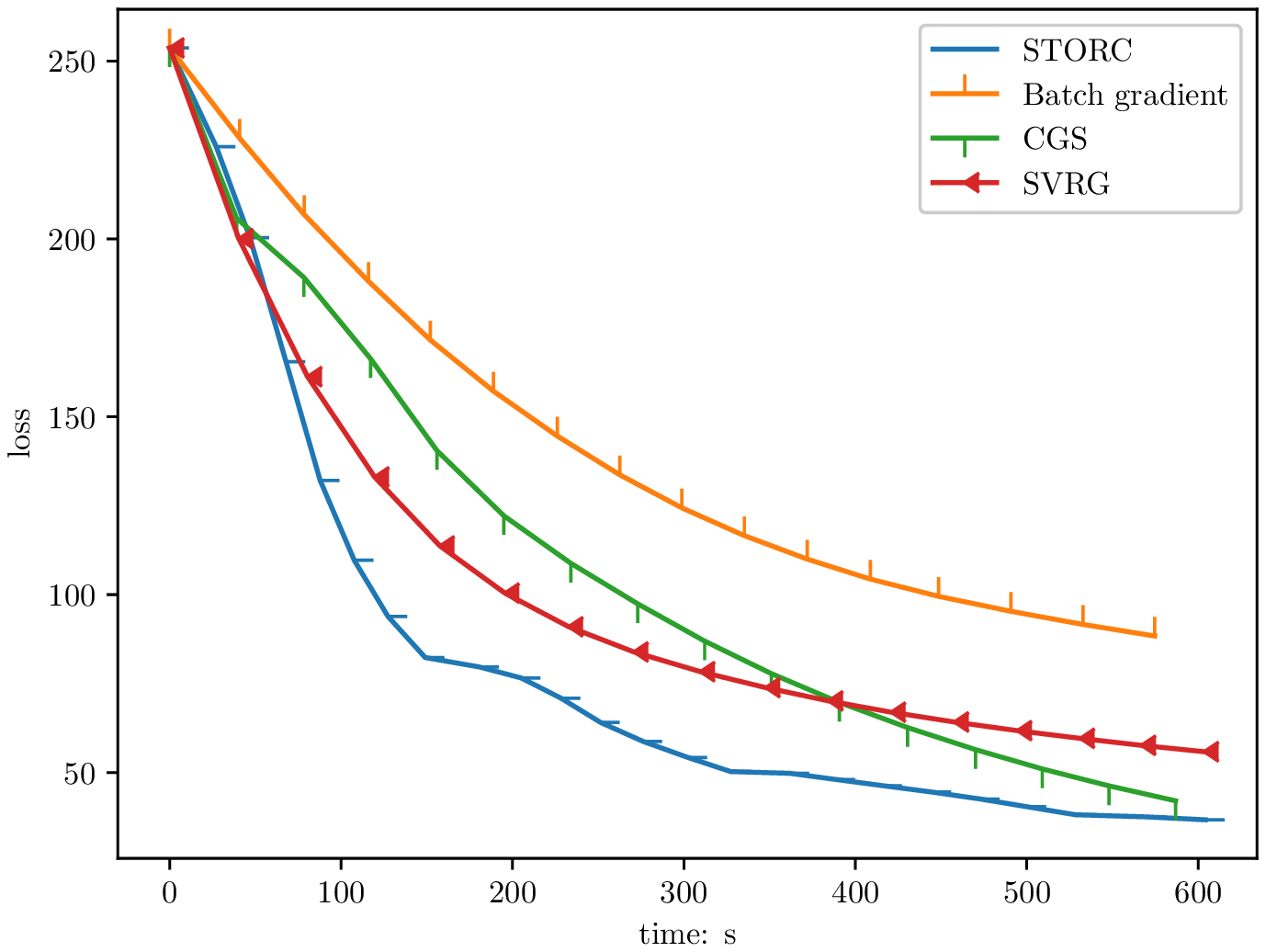}
\caption{$d=250, L/\hat{\sigma} = 1000$}
\end{subfigure}
\begin{subfigure}{.32\textwidth}
\includegraphics[width=\linewidth]{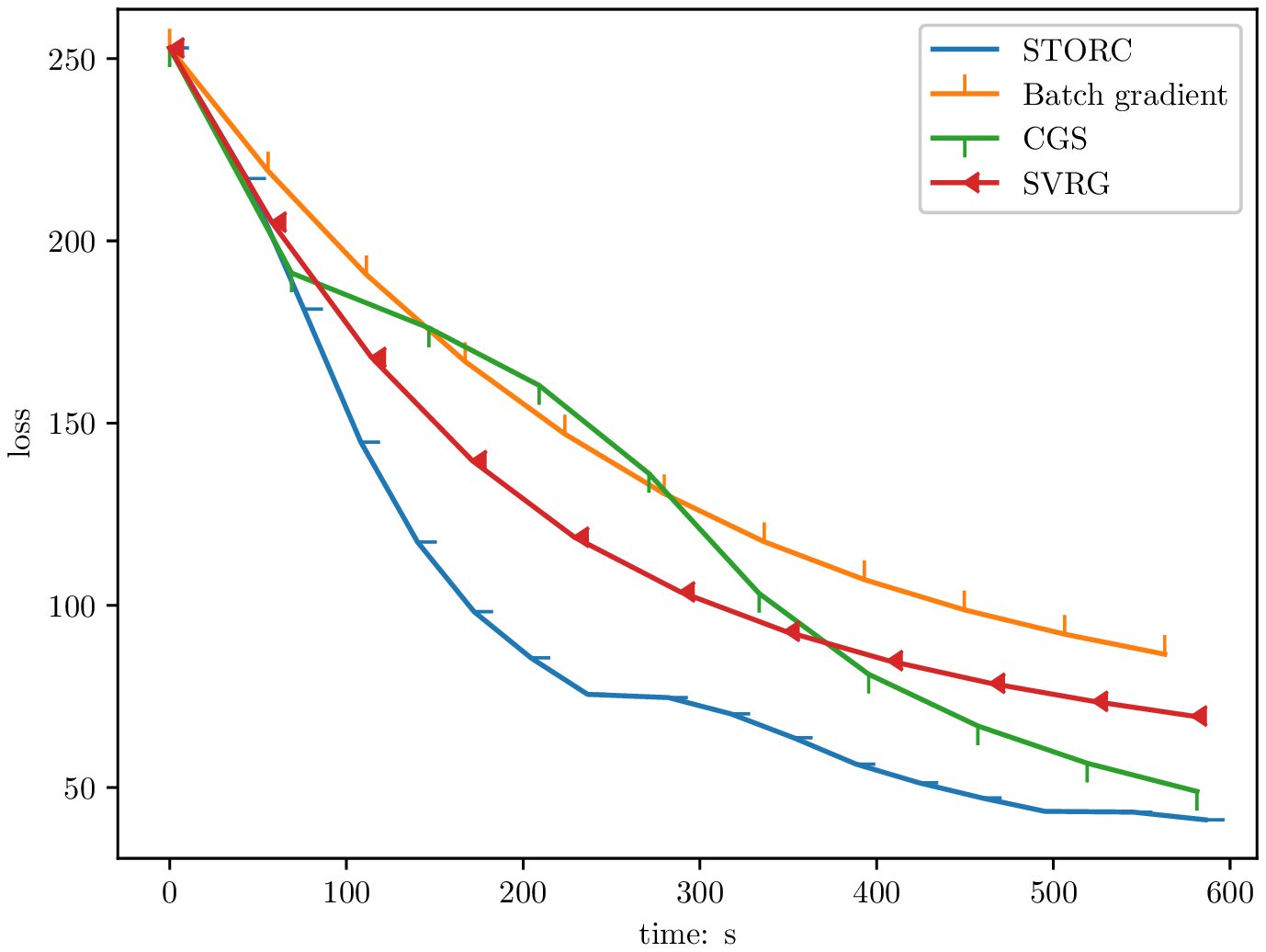}
\caption{$d=300, L/\hat{\sigma} = 1000$}
\end{subfigure}
\begin{subfigure}{.32\textwidth}
\includegraphics[width=\linewidth]{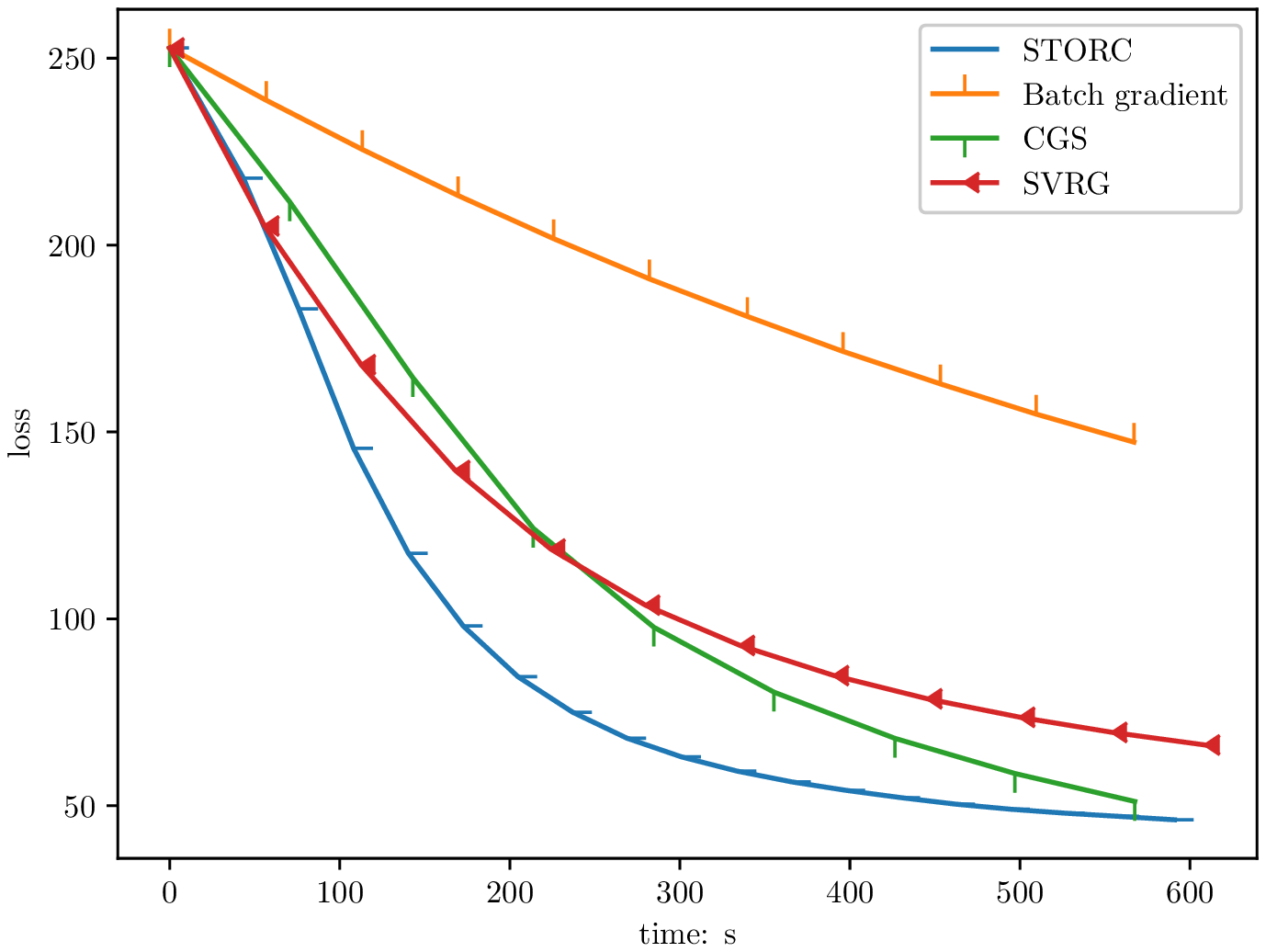}
\caption{$d=300, L/\hat{\sigma} = 10000$} \label{ill_cond}
\end{subfigure}
\caption{Plot of convex empirical loss $L_n(\Theta^t)$ versus computation time. }
\label{fig:convex_plot}
\end{figure}

We compare the performance of CGS and STORC with batch projected gradient descent and SVRG algorithm on matrix regression problem (\ref{matrix_regression}).
For the convex loss problem  we generate data $y_i = \inner{X_i}{\Theta^{\star}} + \epsilon_i$, here $\Theta^{\star}$ is a matrix with rank $r =5 $  with
$\norm{\Theta^{\star}}_*  = 50$. We sample $\vect(X_i) \sim N(0,\Sigma_x)$ with $\Sigma_x$ being diagnal matrix in $\mathbb{R}^{d^2 \times d^2}$
, and every diagnal entry is $\hat{\sigma} = 1$ except the first one being $L$. We set $L =1000, 10000$ to illustrate the impact of condition number performance.
Finally we  generate  $n = \alpha  r d$ samples with  $\alpha = 10$ and set $\rho = 50$.

Figure \ref{fig:convex_plot} reports the simulation result for the convex loss problem.
For batch algorithm CGS and projected gradient descent, the computation time per iteration is dominated by  gradient evaluation.
 As our theorem has predicted, when the required precision
is not too small, the CGS algorithm outperfoms  projected gradient descent due to its acceleration
in terms of gradient evaluation complexity (square root dependence on condition number). This is even more significant when we have ill-conditioned problem as in Figure \ref{ill_cond}.
Both  STORC and and SVRG outperform their batch counterparts by saving gradient evaluations significantly per iteration.
Since STORC performs LO which only requires computing leading singular vectors, it outperforms SVRG which involves full
SVD computation at each iteration.
In fact we observe that SVRG could be even outperformed by CGS, which further emphasizes the importance of replacing full projection by linear optimization oracle.

\begin{figure}
\centering
\begin{subfigure}{.33\textwidth}
\includegraphics[width=\linewidth]{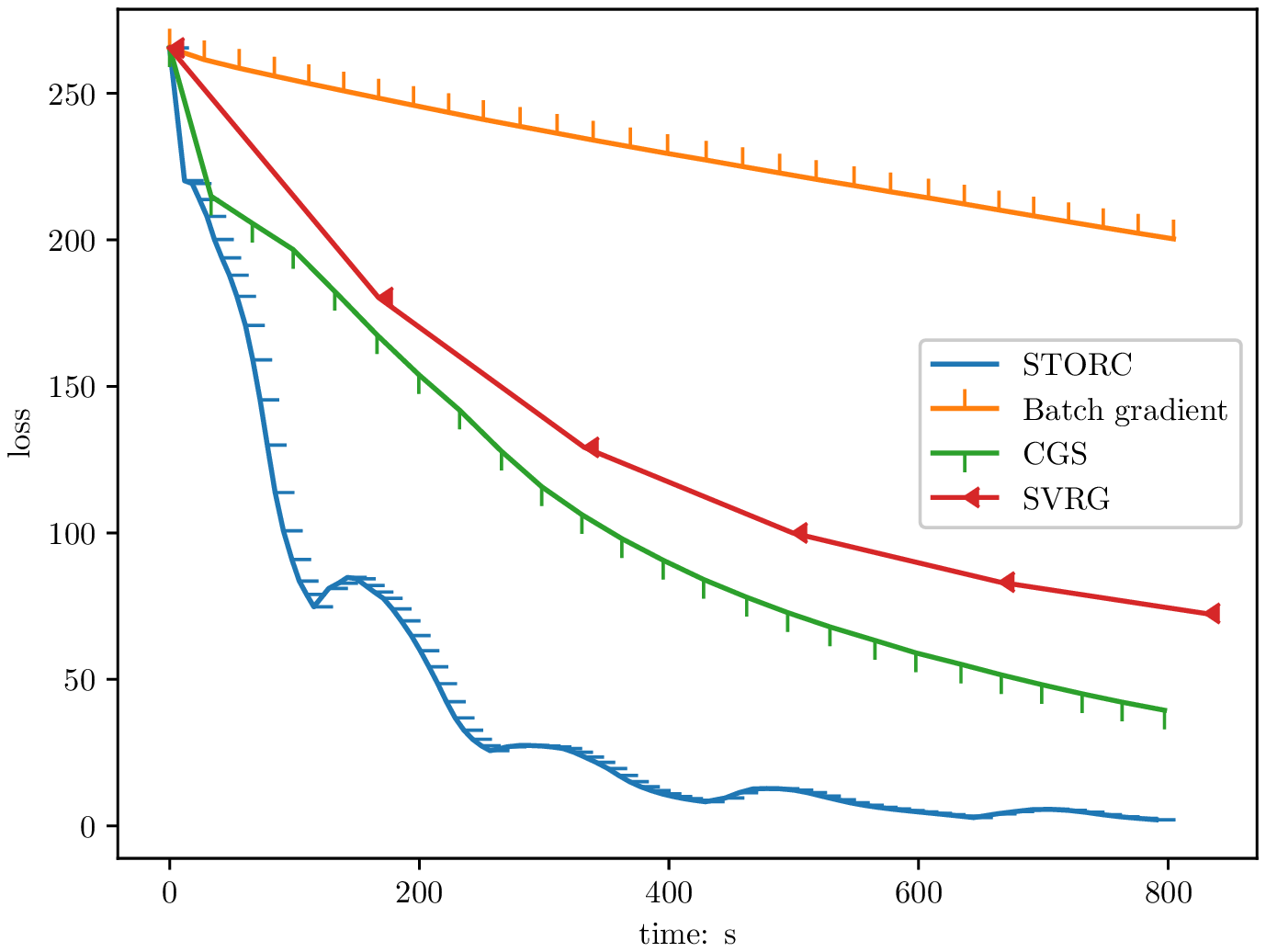}
\caption{$d=250, r=5 $}
\end{subfigure}%
\begin{subfigure}{.33\textwidth}
\includegraphics[width=\linewidth]{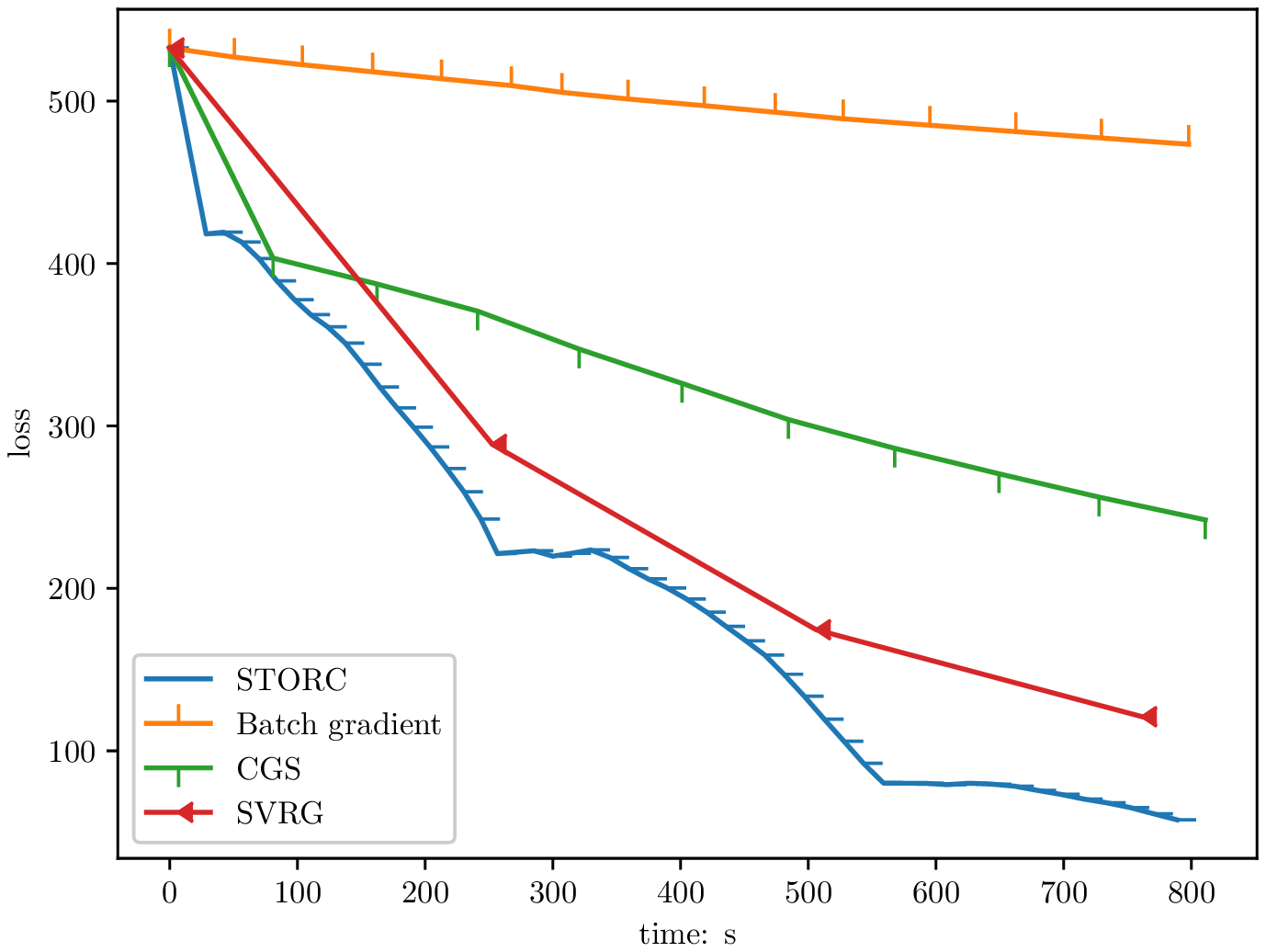}
\caption{$d=250, r=10 $}
\end{subfigure}%
\begin{subfigure}{.33\textwidth}
\includegraphics[width=\linewidth]{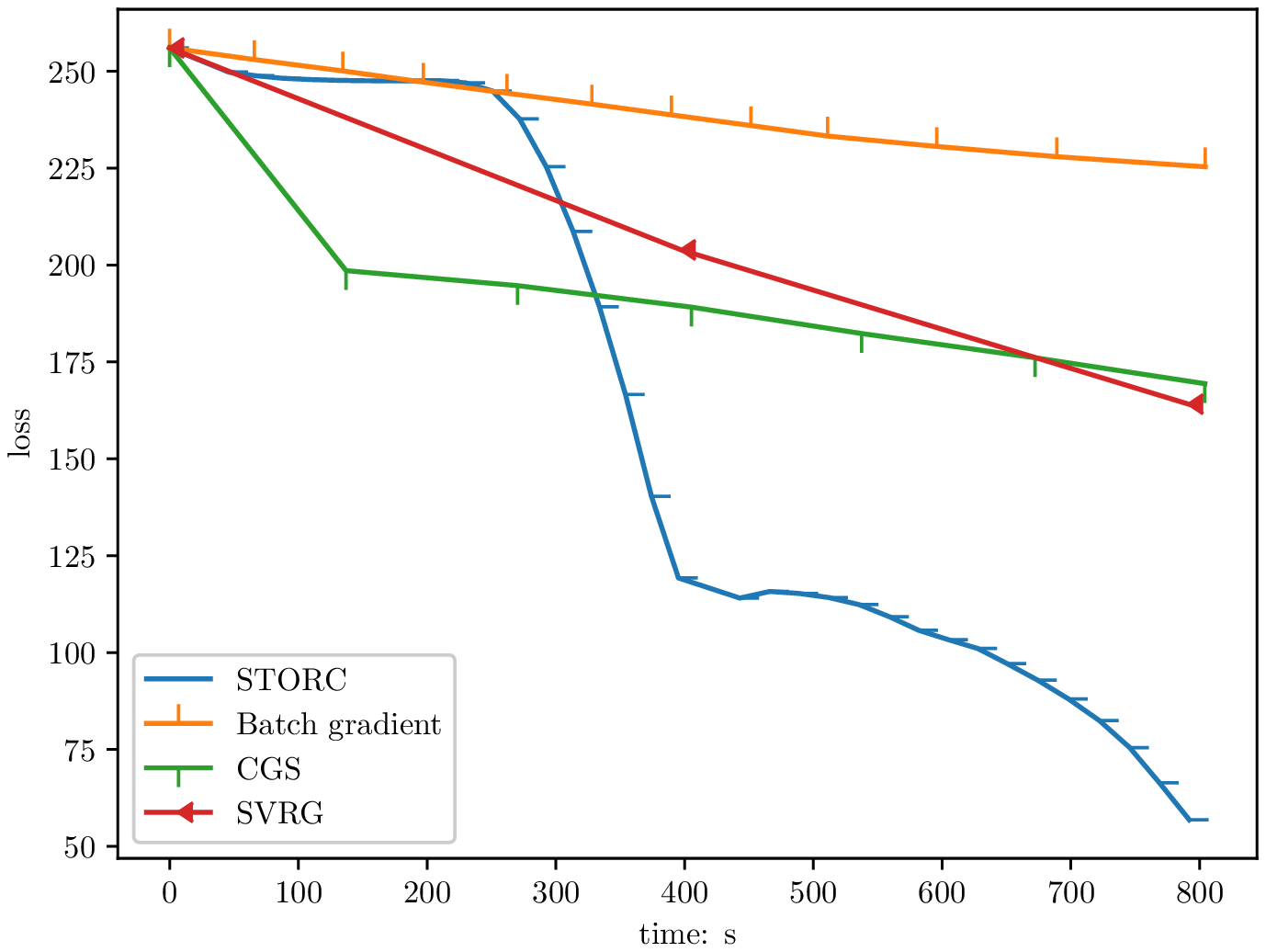}
\caption{$d=300, r=5$}
\end{subfigure}%
\caption{Plot of non-convex empirical loss $L_n(\Theta^t)$ versus computation time. }
\label{fig:nonconvex_plot}
\end{figure}

For non-convex loss problem, we generate $y_i = \inner{X_i}{\Theta^{\star}} + \epsilon_i$ in the same way as the convex loss, where $\hat{\sigma}=1$ and $L = 1000$.
Instead of observing $X_i$ we only observe $Z_i = X_i + W_i$ where $\vect(W_i) \sim N(0, 0.1 \cdot \hat{\sigma}I_{d^2 } )$ and independent of $X_i$. We generate sample of size $n = \alpha rd$
with $\alpha =10$ and set $\rho = 50$.
Figure \ref{fig:nonconvex_plot} reports the simulation results for the non-convex loss. Again, CGS outperforms batch gradient descent by its accelerated performance in terms of gradient evaluation complexity.
Notice that $\sigma_{max}(\Sigma_w) \leqslant \sigma_{min}(\Sigma_x)$ which by our theorem implies that non-convexity essentially causes no extra
computational overhead for STORC algorithm. As predicted by our theoretical result, STORC  outperforms SVRG by replacing projection step  with the much cheaper linear oracle.

\section{Conclusions}
In this paper we show that batch and stochastic variants of Frank-Wolfe algorithm, namely CGS and STORC algorithm
can be used to solve high dimensional statistical estimation problem efficiently, especially when the projection step in the gradient descent type algorithms is computationally hard. While the efficient gradient complexity result for CGS and STORC has been established in literature,
such result  requires  the objective function $f(x) = \frac{1}{n} \sum_{i=1}^n f_i(x)$ being strongly convex and individual
$f_i(x)$ being convex. In this paper we relax these restrictive assumptions that are hardly satisfied in high dimensional-statistics, by
restricted strong convexity that holds true in various statistical mode and   show that the same gradient evaluation complexity
could be maintained under this more general condition.

\newpage
\nocite{*}
\bibliographystyle{elsarticle-harv}
\bibliography{projection_free}

\newpage
\appendix
\rule{\textwidth}{4px}\\
\begin{center}
\textbf{\LARGE Supplemental Material\\
\hfill \\
\Large Projection-Free Algorithms in Statistical Estimation}
\end{center}
\rule{\textwidth}{1px}\\

\section{Proof overview}
In this section we provide a roadmap which we will following in establishing the complexity results presented in the paper.

It is convenient to note that an
$L$-smooth function $f$ satisfies the following properties which is useful in our proof:
\begin{align}\label{smooth-prop}
  \displaystyle
  f(\lambda x +  (1-\lambda) y)   \geqslant
   \lambda f(x) + (1-\lambda) f(y) -\frac{L}{2}\lambda(1-\lambda) ||x-y||^2
\end{align}
for $\lambda \in [0,1]$. If $f$ is additionally convex, then:
\begin{align}\label{convex-smooth}
\norm{\nabla f(x) - \nabla f(y)}^2 \leqslant 2L \left (f(y) -f(x) -\inner{\nabla f(x)}{y-x} \right).
\end{align}

For simplicity of exposition, we emphasize the our proof scheme for convex case here.
The basic idea of our proof is to re-analyze the convergence property of the aforementioned  algorithms, while the strong convexity is replaced by the
restricted strong convexity. More precisely, let $\hat{\Delta}_t = \theta_t -\hat{\theta}$, the first important lemma says that $\varphi(\hat{\Delta}_t)$ in fact
is close to $\norm{\hat{\Delta}_t}$.

\begin{lemma}\label{control_err}
In problem (\ref{prob_formulation}), if $\varphi(\hat{\theta})=\rho$, then we have $\hat{\Delta}_t$ belongs to the set:
\begin{align}\label{reg_bound}
  \mathbb{S}(\mathcal{M, \overline{M}}) = \{\Delta:\, \varphi(\Delta)  \leqslant  2 \phi(\mathcal{\overline{M}}) \norm{\Delta} + 2 \varphi(\Pi_{\mathcal{M}}^{\perp}(\theta^{\star}))
    +2 \varphi(\Delta^{\star}) + 2\phi(\overline{\mathcal{M}})\norm{\Delta^{\star}} \}.
\end{align}
\end{lemma}

Now by combining (\ref{reg_bound}) with the definition of restricted strong convexity, we have the following structure of pseudo strong convexity of  the objective function that \textit{holds at the global optimum}:
\begin{align}\label{pseudo-sc}
 f(\theta)  - f(\hat{\theta})  & \geqslant \inner{\nabla f(\hat{\theta})}{\theta -\hat{\theta}} + \frac{\sigma}{2}\norm{\theta - \hat{\theta}}^2 - \tau_{\sigma} \varphi(\theta-\hat{\theta})^2  \nonumber \\
 & \geqslant \frac{\hat{\sigma}}{2}\norm{\theta-\hat{\theta}}^2 -  \hat{\sigma} \epsilon_{stat}^2,
\end{align}
where the last inequality combines (\ref{reg_bound}), Cauchy-Schwarz inequality and the definition of $\hat{\sigma}, \epsilon_{stat}^2$.

To analyze the convergence of CGS and STORC, we first present a general convergence result of inner iteration. This theorem covers both the deterministic case as in CGS and the stochastic case as in STORC.
\begin{theorem}\label{cgs_convergence}
  Fix outer iteration $t$, if we have for every inner iteration $i$ in CGS and STORC that $\mathbb{E} \left[ \norm{\nabla_i -\nabla f(z_i)}^2 \right ] \leqslant \sigma_i^2 $, then we have:
  \begin{align}\label{gen_convergence}
    \mathbb{E}\left [f(y_k)- f( \hat{\theta}) \right] \leqslant  \beta_1 \Gamma_k \norm{x_0-\hat{\theta}}^2
     + \Gamma_k \sum_{i=1}^k \left [\frac{\eta_{t,i}}{\Gamma_i} + \frac{\gamma_i \sigma_i^2}{2\Gamma_i (\beta_i - L\gamma_i)} \right]
  \end{align}
  where we have defined $\Gamma_i$ as:
  \begin{equation}
    \Gamma_i = \begin{cases}
    1 &  i=1; \\
    (1-\gamma_i) \Gamma_{i-1} & i\geqslant 2.
  \end{cases}
\end{equation}
\end{theorem}
Notice that for CGS algorithm which corresponds to the deterministic case, we have $\sigma_i=0$ for any iteration $i$. By specifying specific parameters in the two algorithms,
we have concrete convergence results for inner iteration, provided we can control $\sigma_i^2$ to decrease in a certain rate. The proof of the following corollary is a simple induction after plugging in all the
parameters into (\ref{gen_convergence}).
\begin{corollary}\label{inner_convergence}
Fix outer iteration $t$, suppose we have $\mathbb{E}\left[\norm{x_0-\hat{\theta}}^2 \right]\leqslant D_t^2 $, then by setting $\gamma_k=\frac{2}{k+1}, \beta_k=\frac{3L}{k}, \eta_{t,k}=\frac{2LD_t^2}{N_t k}$,
if we can control $\sigma_i^2$ such that $\sigma_i^2 \leqslant \frac{L^2D_t^2}{N_t(i+1)^2}$ for all $i \leqslant k$, then we have:
\begin{align}
  \mathbb{E} \left[f(y_k)-f(\hat{\theta}) \right]\leqslant \frac{8LD_t^2}{k(k+1)}.
\end{align}
\end{corollary}
Again, since $\sigma_i=0$ for CGS algorithm, the claim of corollary follows immediately once we have specified appropriate parameters. The main challenge of our proof, is to control $\sigma_i^2$ as prescribed by corollary
for STORC algorithm.

We now explain why Corollary \ref{inner_convergence} almost completes our argument: if we can control $\sigma_i^2 \leqslant \frac{L^2D_t^2}{N_t(i+1)^2} $ for all $i \leqslant N_t$,
then by this corollary we have:
\begin{align}
  \mathbb{E} [f(\theta_t)-f(\hat{\theta}) ]& =  \mathbb{E} [f(y_{N_t})-f(\hat{\theta}) ] \nonumber \\
  & \leqslant \mathbb{E} [\frac{8L\norm{x_0-\hat{\theta}}^2}{N_t(N_t+1)} ]\nonumber \\
  & \leqslant \mathbb{E} [\frac{16\hat{u}(f(\theta_{t-1})-f(\hat{\theta}))}{N_t(N_t+1)} ],
  \label{converge_recursion}
\end{align}
where the first inequality uses $D_t = \norm{x_0-\hat{\theta}}$, and in the second inequality we use the definition of condition number, the pseudo strong convexity (\ref{pseudo-sc}) and the
assumption that we have not achieved the statistical precision yet. Hence by specifying appropriate $N_t$ as in CGS and STORC, we establish the convergence for outer iteration is
\begin{equation}
  \mathbb{E} [f(\theta_t)-f(\hat{\theta}) ] \leqslant \frac{1}{2} \mathbb{E} [f(\theta_{t-1})-f(\hat{\theta}) ].
\end{equation}
With this simple recursion for outer iteration and a careful summation of calls to gradient evaluation and linear optimization per iteration, the claim for CGS and STORC follows immediately,
we leave the details of the proof in the rest of supplemental material.

\section{Proof of Lemma \ref{rsc_lemma}}
\begin{proof}
We have first
\begin{align}
  \mathcal{E}_n (\Delta) & = L_n(\theta + \Delta) - L_n(\theta) - \inner{\nabla L_n(\theta)}{\Delta}\nonumber  \\
  & =  \vect(\Delta)^T \hat{\Gamma} \vect(\Delta)
\end{align}
Define $\zeta(\Sigma) = \sup_{\norm{u}_2=1, \norm{v}_2=1} u^T \Sigma v$.
For convex loss: We have $\mathcal{E}_n (\Delta) = \vect(\Delta)^T (\frac{1}{n} X^T X) \vect(\Delta)$,
by Lemma 7 of \cite{fast_global} we have there exists universl postive constants $c_0, c_1$ such that:
\begin{align}
\mathcal{E}_n(\Delta) \geqslant \frac{1}{2} \lambda_{min}(\Sigma_x) \norm{\Delta}_F^2 - c_1 \zeta(\Sigma_x) \frac{d}{n} \norm{\Delta}_{nuc}^2
\end{align}
with probability at least $1- \exp(-c_0 n)$, which gives result for convex case.

For nonconvex loss, we have $\hat{\Gamma} = \frac{1}{n} Z^T Z - Z_w$, hence $\mathcal{E}_n (\Delta) = \vect(\Delta)^T (\frac{1}{n} Z^T Z) \vect(\Delta) - \vect(\Delta)^T \Sigma_w \vect(\Delta)  = \mathcal{F}(\Delta) - \vect(\Delta)^T \Sigma_w \vect(\Delta)$.
We have that with probability at least $1- \exp(-c_0 n)$,
\begin{align}
  \mathcal{F}_n(\Delta) \geqslant \frac{1}{2} \lambda_{min}(\Sigma_x) \norm{\Delta}_F^2 - c_1 \zeta(\Sigma_z) \frac{d}{n} \norm{\Delta}_{nuc}^2
\end{align}
Now uses assumption that $\lambda_{max}(\Sigma_w) \leqslant \frac{1}{4} \lambda_{min}(\Sigma_x)$ we obtain result immediately.
\end{proof}

\section{Proof of Lemma \ref{control_err}}
\begin{proof}
  First by notice that $\varphi(\hat{\theta})=\rho$, we have $\varphi(\theta_t)\leqslant \varphi(\hat{\theta})$, then by triangle inequality we have
  \begin{align}\label{ub_1}
    \varphi(\theta_t) \leqslant \varphi(\theta^{\star})+\varphi(\Delta^{\star}) \leqslant \varphi \left (\Pi_{\mathcal{M}}(\theta^{\star}) \right ) + \varphi \left (\Pi_{\mathcal{M}^{\perp}}(\theta^{\star})  \right ) + \varphi(\Delta^{\star})
  \end{align}
  Now we lower bound the left hand side using decomposibility of regularization, denote $\Delta_t = \theta_t - \theta^{\star}$:
  \begin{align}\label{lb_1}
    \varphi(\theta_t) & = \varphi \left( \Pi_{\mathcal{M}}(\theta^{\star}) + \Pi_{\mathcal{M}^{\perp}}(\theta^{\star}) + \Pi_{\mathcal{\overline{M}}}(\Delta_t) + \Pi_{\mathcal{\overline{M}}^{\perp}}(\Delta_t) \right ) \nonumber \\
    & \geqslant \varphi \left (\Pi_{\mathcal{M}}(\theta^{\star})+ \Pi_{\mathcal{\overline{M}}^{\perp}}(\Delta_t) \right ) - \varphi \left (\Pi_{\mathcal{M}^{\perp}}(\theta^{\star}) + \Pi_{\mathcal{\overline{M}}}(\Delta_t)\right) \nonumber \\
    & \geqslant \varphi \left (\Pi_{\mathcal{M}}(\theta^{\star})+ \Pi_{\mathcal{\overline{M}}^{\perp}}(\Delta_t) \right ) -\varphi \left (\Pi_{\mathcal{M}^{\perp}}(\theta^{\star})  \right ) -\varphi \left (\Pi_{\mathcal{\overline{M}}}(\Delta_t)\right)  \nonumber \\
    & = \varphi \left (\Pi_{\mathcal{M}}(\theta^{\star}) \right ) +\varphi \left (\Pi_{\mathcal{\overline{M}}^{\perp}}(\Delta_t) \right ) -\varphi \left (\Pi_{\mathcal{M}^{\perp}}(\theta^{\star})  \right ) -\varphi \left (\Pi_{\mathcal{\overline{M}}}(\Delta_t)\right)
  \end{align}
Now by combing (\ref{ub_1}) and (\ref{lb_1}) we have:
\begin{align}
  \varphi \left (\Pi_{\mathcal{\overline{M}}^{\perp}}(\Delta_t) \right ) \leqslant  \varphi \left (\Pi_{\mathcal{\overline{M}}}(\Delta_t)\right) + 2\varphi \left (\Pi_{\mathcal{M}^{\perp}}(\theta^{\star})  \right ) + \varphi(\Delta^{\star})
\end{align}
which in turn by triangle inquality of $\varphi(\cdot)$ we have:
\begin{align}\label{ub_2}
 \varphi \left ( \Delta_t \right) \leqslant 2\varphi \left (\Pi_{\mathcal{\overline{M}}}(\Delta_t)\right) + 2\varphi \left (\Pi_{\mathcal{M}^{\perp}}(\theta^{\star})  \right ) + \varphi(\Delta^{\star})
\end{align}
Finally we have:
\begin{align}
  \varphi \left ( \hat{\Delta_t} \right)  & \leqslant \varphi \left ( \Delta_t \right) +  \varphi(\Delta^{\star})\nonumber  \\
  & \leqslant 2\varphi \left (\Pi_{\mathcal{\overline{M}}}(\Delta_t)\right) + 2\varphi \left (\Pi_{\mathcal{M}^{\perp}}(\theta^{\star})  \right ) + 2\varphi(\Delta^{\star}) \nonumber \\
  & \leqslant 2\phi(\mathcal{\overline{M}}) \norm{\Pi_{\mathcal{\overline{M}}}(\Delta_t)} + 2\varphi \left (\Pi_{\mathcal{M}^{\perp}}(\theta^{\star})  \right ) + 2\varphi(\Delta^{\star}) \nonumber \\
  & \leqslant 2\phi(\mathcal{\overline{M}}) \norm{\Delta_t} + 2\varphi \left (\Pi_{\mathcal{M}^{\perp}}(\theta^{\star})  \right ) + 2\varphi(\Delta^{\star}) \nonumber \\
  & \leqslant 2\phi(\mathcal{\overline{M}}) \norm{\hat{\Delta_t}} +  2\phi(\mathcal{\overline{M}}) \norm{\hat{\Delta^{\star}}} + 2\varphi \left (\Pi_{\mathcal{M}^{\perp}}(\theta^{\star})  \right ) + 2\varphi(\Delta^{\star})
\end{align}
where the first inequality uses bound (\ref{ub_2}), the second equality uses definition of $ \phi(\mathcal{\overline{M}})$, the third inequality uses non-expansiveness of projection and final inequality uses triangle inequality.
\end{proof}

\section{Proof of Theorem \ref{cgs_convergence} and Corollary \ref{inner_convergence}}
\subsection{Proof of Theorem \ref{cgs_convergence}}
The proof of Theorem \ref{cgs_convergence} is essentially centered around the analysis nesterov's acceleration scheme, yet exact projection relaced by approximate projection with
approximate precision controlled by wolfe gap. We present convergence result for stochastic case, which reduce to determinstic case (CGS) with $\sigma_i=0$.
\begin{proof}
Let us fix outer iteration $t$. We first present an important inequality that characterize approximate solution:
\begin{align}\label{proj_lemma}
  \frac{1}{\beta_k} \inner{\nabla_k}{x_k} + \frac{1}{2}\norm{x_k-x}^2 \leqslant \frac{1}{\beta_k}\inner{\nabla_k}{x}+ \frac{1}{2} \norm{x_{k-1}-x}^2 -\frac{1}{2}\norm{x_k-x_{k-1}}^2 +\frac{\eta_{t,k}}{\beta_k}
\end{align}
We note that (\ref{proj_lemma}) comes from wolfe gap condition that $x_k$ satisfies: $\inner{\nabla_k + \beta_k(x_k-x_{k-1})}{x_k -x }\leqslant \eta_{t,k}$ with some algebraic manipulation.
Now by convexity of $f$, definition of $z_k,x_k,y_k$ and smoothness we have:
\begin{align}\label{ub_3}
f(y_k) &\leqslant f(z_k) + \inner{\nabla f(z_k)}{y_k - z_k} +\frac{L}{2}\norm{y_k-z_k}^2 \nonumber \\
& = (1-\gamma_k) \left [ f(z_k) + \inner{\nabla f(z_k)}{y_{k-1} - z_k}\right] + \gamma_k \left [ f(z_k + \inner{\nabla f(z_k)}{x_k - z_k}\right] + \frac{L\gamma_k^2}{2}\norm{x_k-x_{k-1}}^2 \nonumber \\
& \leqslant (1-\gamma_k) f(y_{k-1}) + \gamma_k \left [ f(z_k) + \inner{\nabla f(z_k)}{x_k - z_k}\right] +\frac{\beta_k \gamma_k}{2}\norm{x_k-x_{k-1}}^2 -\frac{\gamma_k}{2}(\beta_k - L\gamma_k) \norm{x_k-x_{k-1}}^2 \nonumber \\
\end{align}
Now let us denote $\delta_k = \nabla f(z_k) -\nabla_k$, and apply projection inequality (\ref{proj_lemma}), we have:
\begin{align}
  f(y_k) & \overset{(\ref{ub_3})}{\leqslant} (1-\gamma_k) f(y_{k-1}) + \gamma_k \left [ f(z_k) + \inner{\nabla_k}{x_k - z_k}\right] +\frac{\beta_k \gamma_k}{2}\norm{x_k-x_k}^2  \nonumber \\
  & \qquad + \gamma_k \inner{\delta_k}{x_k-z_k}  -\frac{\gamma_k}{2}(\beta_k - L\gamma_k) \norm{x_k-x_{k-1}}^2\nonumber \\
  & \overset{(\ref{proj_lemma})}{\leqslant} (1-\gamma_k) f(y_{k-1}) + \gamma_k \left[f(z_k) + \inner{\nabla_k}{x-z_k }\right] +  \frac{\gamma_k \beta_k}{2} \left [\norm{x-x_{k-1}}^2 - \norm{x-x_k}^2\right]  \nonumber \\
  & \qquad + \eta_{t,k} \gamma_k  +  \gamma_k \inner{\delta_k}{x_k-z_k} -\frac{\gamma_k}{2}(\beta_k - L\gamma_k) \norm{x_k-x_{k-1}}^2 \nonumber \\
  & =  (1-\gamma_k) f(y_{k-1}) + \gamma_k \left[f(z_k) + \inner{\nabla f(z_k)}{x-z_k }\right] + \frac{\gamma_k \beta_k}{2} \left [\norm{x-x_{k-1}}^2 - \norm{x-x_k}^2\right]  \nonumber \\
  & \qquad + \eta_{t,k} \gamma_k  +  \gamma_k \inner{\delta_k}{x_k-x} -  \frac{\gamma_k}{2}(\beta_k - L\gamma_k) \norm{x_k-x_{k-1}}^2 \label{cgs_convexity} \\
  & \leqslant (1-\gamma_k) f(y_{k-1}) + \gamma_k f(x) + \frac{\gamma_k \beta_k}{2} \left [\norm{x-x_{k-1}}^2 - \norm{x-x_k}^2\right]   \nonumber  \\
  & \qquad + \eta_{t,k} \gamma_k  +  \gamma_k \inner{\delta_k}{x_{k-1}-x} + \gamma_k \inner{\delta_k}{x_k-x_{k-1}} - \frac{\gamma_k}{2}(\beta_k - L\gamma_k) \norm{x_k-x_{k-1}}^2\nonumber \\
  & \leqslant (1-\gamma_k) f(y_{k-1}) + \gamma_k f(x) + \frac{\gamma_k \beta_k}{2} \left [\norm{x-x_{k-1}}^2 - \norm{x-x_k}^2\right]  \nonumber \\
  & \qquad + \eta_{t,k} \gamma_k  +  \gamma_k \inner{\delta_k}{x_{k-1}-x} + \gamma_k \norm{\delta_k}\norm{x_k-x_{k-1}} - \frac{\gamma_k}{2}(\beta_k - L\gamma_k) \norm{x_k-x_{k-1}}^2\nonumber
\end{align}
By applying basic inequality $bt-\frac{at^2}{2}\leqslant \frac{b^2}{2a}$ to the last line of previous inequality we have:
\begin{align}
  f(y_k) & \leqslant (1-\gamma_k) f(y_{k-1}) + \gamma_k f(x) + \frac{\gamma_k \beta_k}{2} \left [\norm{x-x_{k-1}}^2 - \norm{x-x_k}^2\right]  \nonumber \\
  & \qquad + \eta_{t,k} \gamma_k  +  \gamma_k \inner{\delta_k}{x_{k-1}-x} + \frac{\norm{\delta_k}^2\gamma_k}{\beta_k-L\gamma_k}
\end{align}
Now take expectation of both sizes with obsevation that $\mathbb{E}[\inner{\delta_k}{x_{k-1}-x}]=0$ and $\mathbb{E}[\norm{\delta_k}^2]= \sigma_k^2$, we have:
\begin{align}
  \mathbb{E}[f(y_k)-f(x)] & \leqslant (1-\gamma_k) \mathbb{E}[f(y_{k-1})-f(x)]  + \frac{\gamma_k \beta_k}{2} \mathbb{E} \left [\norm{x-x_{k-1}}^2 - \norm{x-x_k}^2\right]  \nonumber \\
  & \qquad + \eta_{t,k} \gamma_k  +   \frac{\sigma_k^2\gamma_k}{\beta_k-L\gamma_k}
\end{align}
Devide both sides by $\Gamma_k$ and by definition of $\Gamma_k$ we have:
\begin{align}\label{ub_4}
  \frac{1}{\Gamma_k} \mathbb{E}[f(y_k)-f(x)] & \leqslant \frac{1}{\Gamma_{k-1}} \mathbb{E}[f(y_{k-1})-f(x)] + \frac{\gamma_k \beta_k}{2\Gamma_k} \mathbb{E} \left [\norm{x-x_{k-1}}^2 - \norm{x-x_k}^2\right]  \nonumber \\
  & \qquad + \frac{\eta_{t,k} \gamma_k}{\Gamma_k}  +   \frac{\sigma_k^2\gamma_k}{\Gamma_k(\beta_k-L\gamma_k)}
\end{align}
Now sum up (\ref{ub_4}) from $i=1$ to $k$ we have:
\begin{align}\label{ub_5}
  \mathbb{E}[f(y_k)-f(x)] & \leqslant \Gamma_k (1-\gamma_1) \mathbb{E}[f(y_0)-f(x)] + \Gamma_k \sum_{i=1}^k  \frac{\gamma_i \beta_i}{2\Gamma_i} \mathbb{E} \left [\norm{x-x_{i-1}}^2 - \norm{x-x_i}^2\right] \nonumber \\
  & \qquad + \Gamma_k \sum_{i=1}^k \frac{\eta_{t,i} \gamma_i}{\Gamma_i} + \Gamma_k \sum_{i=1}^k \frac{\sigma_i^2\gamma_i}{\Gamma_i(\beta_i-L\gamma_i)} \nonumber \\
  & = \Gamma_k (1-\gamma_1) \mathbb{E}[f(y_0)-f(x)] \nonumber \\
  & \qquad + \Gamma_k \left ( \frac{\gamma_1 \beta_1}{\Gamma_1} \mathbb{E}\norm{x-x_0}^2 + \sum_{i=2}^k (\frac{\gamma_i \beta_i}{2\Gamma_i} -\frac{\gamma_{i-1} \beta_{i-1}}{2\Gamma_{i-1}})\mathbb{E}\norm{x-x_i}^2 - \frac{\gamma_k\beta_k}{2\Gamma_k}\mathbb{E} \norm{x-x_k}^2  \right ) \nonumber \\
  & \qquad + \Gamma_k \sum_{i=1}^k \frac{\eta_{t,i} \gamma_i}{\Gamma_i} + \Gamma_k \sum_{i=1}^k \frac{\sigma_i^2\gamma_i}{\Gamma_i(\beta_i-L\gamma_i)} \nonumber \nonumber \\
  & \leqslant \Gamma_k \beta_1 \mathbb{E}\norm {x-x_0}^2  + \Gamma_k \sum_{i=1}^k \frac{\eta_{t,i} \gamma_i}{\Gamma_i} + \Gamma_k \sum_{i=1}^k \frac{\sigma_i^2\gamma_i}{\Gamma_i(\beta_i-L\gamma_i)}
\end{align}
where (\ref{ub_5}) uses our parameter choice such that $\frac{\gamma_i \beta_i}{\Gamma_i} \leqslant \frac{\gamma_{i-1} \beta_{i-1}}{\Gamma_{i-1}} $ and definition $\gamma_1=1$.
\end{proof}

\subsection{Proof of Corollary \ref{inner_convergence}}
The proof of this corollary is simply plug in our parameter choice in the statement, and do a simple induction, we omit the details here.

\section{Proof of Corollary \ref{batch_cor} and Corollary \ref{stochastic_cor}}
\subsection{Proof of Corollary \ref{batch_cor}}
\begin{proof}
By Lemma \ref{rsc_lemma} we have that $\tau_{\sigma} = \frac{\xi(\Sigma_x)d}{n}$, next we choose appropriate subspace
pair $(\mathcal{M}, \overline{\mathcal{M}}^{\perp})$ to control the tradeoff between denominator and nominator in the definition of $\epsilon_{stat}^2 = \tau_{\sigma}(\varphi(\Pi_{\mathcal{M}^{\perp}}(\theta^{\star}))+2\varphi(\Delta^{\star})+\phi(\overline{\mathcal{M}})\norm{\Delta^{\star}})^2/\hat{\sigma} $.

Let $\Theta^{\star} = UDV^T$ be singular value decomposition of ground truth with rank $r$, let $U^r$ and $V^r$ be first $r$ columns of $U$ and $V$, define:
\begin{align*}
  \mathcal{M} & = \{ \Theta \in \mathbb{R}^{d \times d} : col(\Theta) \subseteq col(U^r), \, col(\Theta^T) \subseteq col(V^r)\} \\
  \mathcal{\overline{M}^{\perp}} & =  \{ \Theta \in \mathbb{R}^{d \times d} : col(\Theta) \subseteq col(U^r)^{\perp}, \, col(\Theta^T) \subseteq col(V^r)^{\perp}\}
\end{align*}
notice now that $\Theta^{\star} \in \mathcal{M}$ and by definition we have $\phi^2(\overline{M}) \leqslant r$, hence we have:
\begin{align}
 \left (\varphi( \Pi_{\mathcal{M^{\perp}}}  (\Theta^{\star}) +2 \varphi (\Delta^{\star}) + \phi(\overline{\mathcal{M}}) \norm{\Delta^{\star}}  \right)^2 \leqslant 12 \varphi^2(\Delta^{\star}) + 3r \norm{\Delta^{\star}}^2 = 12 \norm{\Delta^{\star}}_{nuc}^2 + 3 r \norm{\Delta^{\star}}_F^2
\end{align}
We use an inequality which will be verified later: $\norm{\Delta^{\star}}_{nuc}^2 \leqslant 4r \norm{\Delta^{\star}}_F^2$, summarizing all above we have:
\begin{align}
  \epsilon_{stat}^2 = \frac{\xi(\Sigma_x) d }{n \hat{\sigma}} \cdot cr \norm{\Delta^{\star}}_F^2
\end{align}
for some constant $c$.

Now $L = \sigma_{max}(\Sigma_x)$, $\hat{\sigma } = \sigma_{min}(\Sigma_x) - \frac{\xi(\Sigma_x) rd}{n}$ follows directy from its definition and Lemma \ref{rsc_lemma}, with scaling $n = \Omega(\frac{\xi(\Sigma_x)r}{d \sigma_{min}(\Sigma_x)})$
we have $\hat{\sigma} = c_1 \sigma_{min}(\Sigma_x)$ for some constant $c_1$. And we also have simple inequality $\norm{X}_F^2 \leqslant \norm{X}_{nuc}^2$ for any matrix, hence $D \leqslant \rho$, plug in
specification of $L, \hat{\sigma}, D$ yields bound on gradient evaluation and linear oracle. Finally the scaling of $n = \Omega( \frac{\xi(\Sigma_x) rd}{\sigma_{\min}(\Sigma_x)})$
would also gives us $\epsilon_{stat}^2 \leqslant c_2 \norm{\Delta^{\star}}_F^2$, which gives us the bound on distance to optimum.

It remains to show $\norm{\Delta^{\star}}_{nuc}^2 \leqslant 4r \norm{\Delta^{\star}}_F^2$, this is very similar to the proof of Lemma \ref{control_err},
using condition that $\varphi(\hat{\theta}) \leqslant \rho \leqslant $ and use triangle inequality we have
 $ \varphi(\hat{\theta}) \leqslant \varphi( \Pi_{\mathcal{M}} (\theta^{\star})) + \varphi(\Pi_{\mathcal{M}^{\perp}} (\theta^{\star})) $.
 Now we lower bound the left hand side.
 \begin{align*}
   \varphi(\hat{\theta}) & = \varphi(\theta^{\star} + \delta^{\star}) \\
   & = \varphi(\Pi_{\mathcal{M}} (\theta^{\star}) + \Pi_{\mathcal{M}^{\perp}} (\theta^{\star}) + \Pi_{\mathcal{\overline{M}}} (\Delta^{\star})  + \Pi_{\mathcal{\overline{M^{\perp}}}} (\Delta^{\star})) \\
   & \geqslant \varphi(\Pi_{\mathcal{M}} (\theta^{\star}) + \Pi_{\mathcal{\overline{M^{\perp}}}} (\Delta^{\star})) - \varphi(\Pi_{\mathcal{M}^{\perp}} (\theta^{\star})) - \varphi( \Pi_{\mathcal{\overline{M}}} (\Delta^{\star})) \\
   & = \varphi(\Pi_{\mathcal{M}} (\theta^{\star})) + \varphi( \Pi_{\mathcal{\overline{M^{\perp}}}} (\Delta^{\star}) ) - \varphi(\Pi_{\mathcal{M}^{\perp}} (\theta^{\star})) - \varphi( \Pi_{\mathcal{\overline{M}}} (\Delta^{\star}))
 \end{align*}
 combined with the upper bound we have:
 \begin{align*}
   \varphi( \Pi_{\mathcal{\overline{M^{\perp}}}} (\Delta^{\star})) \leqslant \varphi( \Pi_{\mathcal{\overline{M}}} (\Delta^{\star})) + 2\varphi(\Pi_{\mathcal{M}^{\perp}} (\theta^{\star}) )
 \end{align*}
 which by triangle inequality implies:
 \begin{align*}
   \varphi(\Delta^{\star}) & \leqslant 2 \varphi( \Pi_{\mathcal{\overline{M}}} (\Delta^{\star})) + 2\varphi(\Pi_{\mathcal{M}^{\perp}} (\theta^{\star}) ) \\
   & \leqslant 2 \phi(\overline{\mathcal{M}}) \norm{\Pi_{\mathcal{\overline{M}}} (\Delta^{\star})} + 2\varphi(\Pi_{\mathcal{M}^{\perp}} (\theta^{\star}) ) \\
   & \leqslant 2 \phi(\overline{\mathcal{M}}) \norm{\Delta^{\star}} + 2\varphi(\Pi_{\mathcal{M}^{\perp}} (\theta^{\star}) )
 \end{align*}
notice now that by out definition of $\mathcal{M}$ in the example we have $\varphi(\Pi_{\mathcal{M}^{\perp}} (\theta^{\star}) ) =0$, specifically we have: $\norm{\Delta^{\star}}_{nuc} \leqslant 2 \phi(\overline{\mathcal{M}}) \norm{\Delta^{\star}}_F $,
now take square on both sides and use the fact that $\phi^2(\overline{\mathcal{M}}) = r$ gives $\norm{\Delta^{\star}}_{nuc}^2 \leqslant 4r \norm{\Delta^{\star}}_F^2$.
\end{proof}

\subsection{Proof of Corollary \ref{stochastic_cor}}
\begin{proof}
This is essentially the same as in proof of corollary \ref{batch_cor},under the scaling $n = \Omega( \frac{\xi(\Sigma_x) dr}{\sigma_{min}(\Sigma_x)})$ we have
 modular some costants, $\epsilon_{stat}^2 = \frac{ \xi(\Sigma_x)dr}{n \hat{\sigma}} \norm{\Delta^{\star}}_F^2, L = \sigma_{max}(\Sigma_x), \hat{\sigma} = \sigma_{\min}(\Sigma_x), D = \rho$
plug in these into theorem \ref{thrm:STORC} yields the result.
\end{proof}

\section{Proof of Theorem \ref{thrm:CGS} and Theorem \ref{thrm:STORC}: Convex Case}
We are now ready to prove our main theorem, as stated before if condition of Corollary \ref{inner_convergence} holds for both CGS and STORC, then we can do this in a unified way.
As we will show in the proof, the parameter specification of $\beta_k, \gamma_k , \eta_{t,k}$ in both CGS and STORC will satisfy condition in the corollary for some $D_t$, the remaining work is to control $\sigma_i^2$.
Since CGS algorithm is determinstic, we now only need to control $\sigma_i^2$ in STORC algorithm, and this is where variance reduction came into play.

\begin{proof}
We will show by induction that
\begin{align}\label{outer_convergence}
\mathbb{E} \left[f(\theta_t) - f(\hat{\theta})\right] \leqslant \frac{\delta_0}{2^{t}}
\end{align}

For base case $t=0$, $\mathbb{E} \left [f(\theta_0) -f(\hat{\theta}) \right] \leqslant \delta_0$ by definition.

\textbf{Analysis for STORC: }Suppose the (\ref{outer_convergence}) holds for all outer iteration before iteration $t$, we now focus on interation $t$. First we upper bound the variance of $\nabla_k$ in STORC algorithm. We claim that:
\begin{align}\label{vr_bound}
  \mathbb{E} [\norm{\nabla f_j(z_k) -\nabla f_j(y_0) + \nabla f(y_0) - \nabla f(z_k)}^2] \leqslant 4L \mathbb{E} \left ([f(z_k) -f(\hat{\theta})] + [f(y_0) - f(\hat{\theta})] \right)
\end{align}
We note that by construction of $\nabla_k$ we have:
\begin{align}\label{vr_bound_2}
\sigma_i^2  & = \frac{1}{m_{t,i}} \mathbb{E} \norm{\nabla f_j(z_i) -\nabla f_j(y_0) + \nabla f(y_0) - \nabla f(z_i)}^2 \nonumber \\
 & \leqslant \frac{4L}{m_{t,i}}  \mathbb{E} \left ([f(z_i) -f(\hat{\theta})] + [f(y_0) - f(\hat{\theta})] \right)
\end{align}

For now we assume (\ref{vr_bound}) holds, and we next show $\sigma_i^2 \leqslant \frac{L^2D_t^2}{N_t(i+1)^2} $ by induction.

For base case $i=1$, notice that $z_0 = y_0 =x_0$, by (\ref{vr_bound_2}) we have:
\begin{align}\label{ub_sig_1}
\sigma_1^2  \leqslant \frac{8L}{m_{t,1}} \mathbb{E} [f(y_0) - f(\hat{\theta})] &  =  \frac{8L}{m_{t,1}} \mathbb{E} [f(\theta_{t-1}) - f(\hat{\theta})]  \leqslant \frac{8L\delta_0}{m_{t,1}2^{t-1}}
\end{align}
Now by pseudo strong convexity (\ref{pseudo-sc}) and assumption that we have not converged to statistical precision, we have:
\begin{align}
  \mathbb{E}  \norm{x_0 -\hat{\theta}}^2  & \leqslant \mathbb{E} \left[\frac{(f(x_0)-f(\hat{\theta})) + \hat{\sigma}\epsilon_{stat}^2}{\hat{\sigma}} \right ]\nonumber \\
  & \leqslant \mathbb{E} \left [\frac{2(f(x_0)-f(\hat{\theta}))}{\hat{\sigma}} \right ]\nonumber \\
  & \leqslant \frac{\delta_0}{2^{t-2}\hat{\sigma}}
\end{align}
Hence we can choose $D_t^2 = \frac{\delta_0}{2^{t-2}\hat{\sigma}}$, combine this with (\ref{ub_sig_1}) we have $\sigma_1^2 \leqslant \frac{4LD_t^2\hat{\sigma}}{m_{t,1}} \leqslant \frac{L^2 D_t^2}{4N_t}$, the proof for $i=1$ completes.

Suppose (\ref{vr_bound}) holds for $i<k$, by (\ref{vr_bound_2}) to control $\sigma_k^2$ we need to upper bound $f(z_k)$. Now by definition of $z_k, x_k, y_k$ and $\gamma_k$ we have:
\begin{align}\label{convex_combi}
  y_{k-1}= \frac{k+1}{2k-1} z_k +\frac{k-2}{2k-1} y_{k-2}
\end{align}
Together with smoothness property (\ref{smooth-prop}) we have:
\begin{align}
 f(y_{k-1}) -f(\hat{\theta})& \geqslant \frac{k+1}{2k-1} \left(f(z_k)-f(\hat{\theta}) \right) + \frac{k-2}{2k-1} \left(f(y_{k-2})-f(\hat{\theta}) \right)  - \frac{L (s+1)(s-2)}{2(2s-1)^2}  \norm {z_k -y_{k-2}}^2 \nonumber \\
 & \geqslant \frac{1}{2} \left(f(z_k) - f(\hat{\theta}) \right) -\frac{L}{2} \norm {y_{k-1} -y_{k-2}}^2 \nonumber \\
\end{align}
Rearrange terms, use cauchy-schwarz inequality and pseudo strong convexity (\ref{pseudo-sc}) we have upper bound for $f(z_k)$:
\begin{align}
 \mathbb{E} \left[f(z_k) -f(\hat{\theta}) \right]& \leqslant 2 \mathbb{E}\left(f(y_{k-1}) - f(\hat{\theta}) \right) + L \mathbb{E}\norm {y_{k-1} -y_{k-2}}^2 \nonumber \\
 & \leqslant 2 \mathbb{E}\left(f(y_{k-1}) - f(\hat{\theta}) \right) + 2L \mathbb{E}(\norm{y_{k-1} -\hat{\theta}}^2 + \norm{y_{k-2}- \hat{\theta}}^2) \nonumber \\
 & \overset{(\ref{pseudo-sc})}{\leqslant} 2\mathbb{E}\left (f(y_{k-1}) - f(\hat{\theta}) \right) + 2L \mathbb{E}\left(\frac{f(y_{k-1})- f(\hat{\theta}) + \hat{\sigma}\epsilon_{stat}^2}{\hat{\sigma}} + \frac{f(y_{k-2})- f(\hat{\theta}) + \hat{\sigma}\epsilon_{stat}^2}{\hat{\sigma}} \right ) \nonumber \\
 & \leqslant 2 \mathbb{E} \left(f(y_{k-1}) - f(\hat{\theta}) \right) + \frac{8L}{\hat{\sigma}} \mathbb{E} \left([f(y_{k-1})- f(\hat{\theta})]+[f(y_{k-2})- f(\hat{\theta})] \right)  \label{precision_assumption} \\
 & \leqslant \frac{2L}{\hat{\sigma}} \mathbb{E} \left(f(y_{k-1}) - f(\hat{\theta}) \right) + \frac{8L}{\hat{\sigma}}\mathbb{E} \left([f(y_{k-1})- f(\hat{\theta})]+[f(y_{k-2})- f(\hat{\theta})] \right) \label{induction_bound}
\end{align}
Note that in (\ref{precision_assumption}) we uses our assumption that we have not converged to statistical precision yet, that is $\text{min} \{ \mathbb{E} [f(y_{k-1})- f(\hat{\theta})], \mathbb{E} [f(y_{k-2})- f(\hat{\theta})] \} \geq \hat{\sigma}\epsilon_{stat}^2$.

Now using (\ref{induction_bound}) we can utilize upper bound for $f(y_{k-1})$ and $f(y_{k-2})$.
Using induction hypothesis that (\ref{vr_bound}) holds for $i<k$, by an application of Corollary \ref{inner_convergence} we know that $\mathbb{E}[f(y_{k-1}) -f(\hat{\theta})] \leqslant \frac{8LD_t^2}{k(k-1)}$ and $\mathbb{E}[f(y_{k-2})-f(\hat{\theta})] \leqslant \frac{8LD_t^2}{(k-2)(k-1)}$, note $\hat{u} = \frac{L}{\hat{\sigma}}$, we have:
\begin{align}
\mathbb{E} [f(z_k)-f(\hat{\theta})] & \leqslant 10\hat{u} \mathbb{E} \left(f(y_{k-1}) - f(\hat{\theta}) \right) + 8\hat{u} \mathbb{E}\left(f(y_{k-2}) - f(\hat{\theta}) \right) \nonumber \\
& \leqslant \frac{80 \hat{u} LD_t^2}{k(k-1)} + \frac{64 \hat{u}LD_t^2}{(k-2)(k-1)} \nonumber \\
& \leqslant \frac{1200 \hat{u} L D_t^2}{(k+1)^2}
\end{align}
Finally, by plugging the above result in (\ref{vr_bound_2}) we have:
\begin{align}
  \sigma_k^2 & \leqslant \frac{4L}{m_{t,k}}  \mathbb{E} \left ([f(z_k) -f(\hat{\theta})] + [f(y_0) - f(\hat{\theta})] \right) \nonumber \\
  & \leqslant \frac{4L}{m_{t,k}}   \left (\frac{1200 \hat{u} L D_t^2}{(k+1)^2} + \frac{\delta_0}{2^{t-1}} \right) \nonumber \\
  & = \frac{4L}{m_{t,k}}   \left (\frac{1200 \hat{u} L D_t^2}{(k+1)^2} + \frac{2LD_t^2}{\hat{u}} \right) \nonumber \\
  & \leqslant \frac{4L}{m_{t,k}}  \left (\frac{1200 \hat{u} L D_t^2}{(k+1)^2} + \frac{64LD_t^2}{(N_t +1)^2} \right) \nonumber \\
  & \leqslant \frac{4L}{m_{t,k}}   \left (\frac{1200 \hat{u} L D_t^2}{(k+1)^2} + \frac{64LD_t^2}{(k+1)^2} \right) \nonumber \\
  & \leqslant \frac{L^2D_t^2}{N_t (k+1)^2}
\end{align}
where the last inequality uses our choice of $m_{t,k}$. We have now completed inductinon for showing $\sigma_k^2 \leqslant \frac{L^2D_t^2}{N_t (k+1)^2}$ for $k=1 \text{ to } N_t$, by choosing $k=N_t$ and use corollary (\ref{inner_convergence}) we have:
\begin{align}
  \mathbb{E} [f(y_{N_t}) -f(\hat{\theta})] & \leqslant \frac{8LD_t^2}{N_t(N_t+1)} \nonumber \\
  & \leqslant \frac{\delta_0}{2^{t}}
\end{align}
where the last inequality comes from our choice of $N_t $. By $f(\theta_t) = f(y_{N_t})$ we have also completed induction for outer iteration.

\textbf{Analysis for CGS: }
Since $\sigma_k^2 =0$ for $k=1 \text{ to } N_t$, choose $D_t^2$ as in STORC algorithm, by corollary \ref{inner_convergence} we immeditely have that $f(\theta_t) - f(\hat{\theta}) \leqslant \frac{8LD_t^2}{N_t(N_t+1)} \leqslant \frac{\delta_0}{2^{t}}$ by our choice of $N_t$.

\textbf{Parameter Specification:} We should note that both CGS and STORC uses exactly the same paramter specification for $\gamma_k=\frac{2}{k+1}, \beta_k=\frac{3L}{k}$, and $\eta_{t,k} = \frac{8L\delta_0}{\hat{\sigma} 2^t N_t k} = \frac{2LD_t^2}{N_t k}$, hence they all meet the
condition in corollary \ref{inner_convergence}.

\textbf{Gradient Complexity and LO Complexity: }
Now we need to sum up calls to gradient evaluation complexity, by (\ref{outer_convergence}) we know that to get $\epsilon$-precision solution we need $T = \log_2(\frac{\delta_0}{\epsilon})$ outer iteration for both CGS and STORC.

For CGS algorithm calls to gradient evalution is bounded by:
\begin{align}
  \sum_{t=1}^T N_t = \mathcal{O}(T \sqrt{\hat{u}}) = \mathcal{O} \left (\sqrt{\frac{L}{\hat{\sigma}}}\log(\frac{\delta_0}{\epsilon}) \right)
\end{align}
it was shown that duality gap of frank-wolfe algorithm decrease with rate $\mathcal{O}(\frac{1}{m})$, and to solve projection approximately
in the algorithm takes $\mathcal{O} \left (\frac{\beta_kD^2}{\eta_{t,k}} \right)$ iteration \cite{revisit_fw_jaggi}, hence we have
calls to the linear oracle bounded by:
\begin{align}
  \sum_{t=1}^T \sum_{k=1}^{N_t} \mathcal{O} \left (\frac{\beta_kD^2}{\eta_{t,k}} +1  \right) &= \sum_{t=1}^T  \mathcal{O} \left (  \frac{N_t^2D^2}{D_t^2} + N_t\right) \nonumber \\
  & = \sum_{t=1}^{T} \mathcal{O}\left (\frac{2^t LD^2}{\delta_0} +N_t\right)\nonumber \\
  & = \mathcal{O} \left (\frac{LD^2}{\epsilon} + \sqrt{\frac{L}{\hat{\sigma}}} \log(\frac{\delta_0}{\epsilon})\right )
\end{align}

For STORC algorithm calls to stochastic gradient evaluation is bounded by:
\begin{align}
  \sum_{t=1}^T (n+ \sum_{k=1}^{N_t} m_{t,k} )& = \sum_{t=1}^T\mathcal{O} \left(n + N_t^2 \hat{u} \right) \nonumber \\
  & = \mathcal{O} \left ((n + (\frac{L}{\hat{\sigma}})^2 ) \log_2(\frac{\delta_0}{\epsilon}) \right)
\end{align}
and similar to CGS, the calls to the linear oracle is bounded by:
\begin{align}
  \sum_{t=1}^T \sum_{k=1}^{N_t} \mathcal{O} \left (\frac{\beta_kD^2}{\eta_{t,k}} +1  \right) &= \sum_{t=1}^T  \mathcal{O} \left (  \frac{N_t^2D^2}{D_t^2} + N_t\right) \nonumber \\
  & = \sum_{t=1}^{T} \mathcal{O}\left (\frac{2^t LD^2}{\delta_0} +N_t\right)\nonumber \\
  & = \mathcal{O} \left (\frac{LD^2}{\epsilon} + \sqrt{\frac{L}{\hat{\sigma}}} \log(\frac{\delta_0}{\epsilon})\right )
\end{align}

\end{proof}

The only remaining part is to show the upper bound for variance of stochastic gradient (\ref{vr_bound}):
\begin{proof}
  \begin{align}
    & \mathbb{E} [\norm{\nabla f_j(z_k) -\nabla f_j(y_0) + \nabla f(y_0) - \nabla f(z_k)}^2] \nonumber \\
    & \leqslant \mathbb{E} [\norm{\nabla f_j(z_k) -\nabla f_j(y_0) )}^2] \nonumber \\
    & = \mathbb{E} [\norm{ (\nabla f_j(z_k) - \nabla f_j(\hat{\theta}) ) - (\nabla f_j(y_0) -\nabla f_j(\hat{\theta}) )}^2] \nonumber \\
    & \leqslant 2 \mathbb{E} \left(\norm{\nabla f_j(z_k) - \nabla f_j(\hat{\theta})}^2 + \norm{\nabla f_j(y_0) -\nabla f_j(\hat{\theta})}^2 \right) \nonumber \\
    & \overset{(\ref{convex-smooth})}{\leqslant} 4L \mathbb{E} \left([f_j(z_k)- f_j(\hat{\theta}) -\inner{\nabla f_j(\hat{\theta})}{z_k-\hat{\theta}}] +[f_j(y_0)- f_j(\hat{\theta}) -\inner{\nabla f_j(\hat{\theta})}{y_0-\hat{\theta}}] \right) \nonumber \\
    & = 4L  \mathbb{E}\left ([f(z_k)- f(\hat{\theta}) -\inner{\nabla f(\hat{\theta})}{z_k-\hat{\theta}}] +f(y_0)- f(\hat{\theta}) -\inner{\nabla f(\hat{\theta})}{y_0-\hat{\theta}}] \right ) \nonumber \\
    & \leqslant 4L \mathbb{E} \left([f(z_k) -f(\hat{\theta})] + [f(y_0) - f(\hat{\theta})] \right)
  \end{align}
  where the last inequality uses optimality of $\hat{\theta}$.
\end{proof}

\section{Proof of Theorem \ref{thrm:CGS} and Theorem \ref{thrm:STORC}: Non-convex Case}
\subsection{Proof of Theorem \ref{thrm:CGS}}
\begin{proof}
  By definition of restricted strong convexity we have:
  \begin{align}
    f(\hat{\theta}) -f(\theta) \leqslant \inner{\nabla f(\theta)}{\hat{\theta}-\theta} +\frac{\hat{\sigma}}{2}\norm{\theta -\hat{\theta}}^2 - \hat{\sigma}\epsilon_{stat}^2
  \end{align}
  Now pluging this bound into (\ref{cgs_convexity}) with $\theta = z_k$ and $x=\hat{\theta}$ we obtain:
  \begin{align}
    f(y_k)  & \leqslant (1-\gamma_k) f(y_{k-1}) + \gamma_k f(\hat{\theta}) + \gamma_k \hat{\sigma}\epsilon_{stat}^2 + \frac{\gamma_k \beta_k}{2} \left [\norm{\hat{\theta}-x_{k-1}}^2 - \norm{\hat{\theta}-x_k}^2\right]   \nonumber  \\
      & \qquad + \eta_{t,k} \gamma_k  +  \gamma_k \inner{\delta_k}{x_{k-1}-\hat{\theta}} + \gamma_k \inner{\delta_k}{x_k-x_{k-1}} - \frac{\gamma_k}{2}(\beta_k - L\gamma_k) \norm{x_k-x_{k-1}}^2\nonumber
  \end{align}
  Following exactly the same argument in the proof of Theorem \ref{cgs_convergence}, we would end up having:
  \begin{align}
    f(y_k) - f(\hat{\theta}) & \leqslant \Gamma_k \beta_1 \mathbb{E} \norm{\hat{\theta}-x_0}^2  + \Gamma_k \sum_{i=1}^k \frac{\eta_{t,i} \gamma_i}{\Gamma_i} +  \Gamma_k \sum_{i=1}^k \frac{\gamma_i \hat{\sigma} \epsilon_{stat}^2}{\Gamma_i}
  \end{align}
  We again will proceed with induction by showing that $f(\theta_t) - f(\hat{\theta}) \leqslant \frac{\delta_0}{2^t}$. The base case is trivial by definition of $\delta_0$. Now suppose the claim holds before iteration $t$,
  then by pseudo strong convexity (\ref{pseudo-sc}) we have:
  \begin{align}
    \mathbb{E}  \norm{x_0 -\hat{\theta}}^2  & \leqslant \mathbb{E} \left[\frac{(f(x_0)-f(\hat{\theta})) + \hat{\sigma}\epsilon_{stat}^2}{\hat{\sigma}} \right ]\nonumber \\
    & \leqslant \mathbb{E} \left [\frac{2(f(x_0)-f(\hat{\theta}))}{\hat{\sigma}} \right ]\nonumber \\
    & \leqslant \frac{\delta_0}{2^{t-2}\hat{\sigma}}
  \end{align}
  Hence we can choose $D_t^2 = \frac{\delta_0}{2^{t-2}\hat{\sigma}}$. Plug in our choice of parameters, and simple calculation yields the following:
  \begin{align}
    f(y_k) - f(\hat{\theta}) \leqslant \frac{8LD_t^2}{k(k+1)} + \hat{\sigma}\epsilon_{stat}^2
  \end{align}
  Note that we have $\frac{8LD_t^2}{k(k+1)} = \frac{32 \delta_0 \hat{u}}{k(k+1)} \geqslant \frac{\delta_0}{2^{t-1}} \geqslant \hat{\sigma} \epsilon_{stat}^2 $ where the last inequality uses the assumption that we have not converged to statistical precision,
  we further obtain:
  \begin{align}\label{inner_convergence_nonconvex}
    f(y_k) - f(\hat{\theta}) \leqslant \frac{16LD_t^2}{k(k+1)}
  \end{align}

  Now by (\ref{inner_convergence_nonconvex}) we obtain:
  \begin{align}
    f(\theta_t) - f(\hat{\theta}) =f(y_{N_t}) - f(\hat{\theta})   \leqslant \frac{16LD_t^2}{k(k+1)} \leqslant \frac{64L \delta_0}{N_t(N_t+1) 2^{t} \hat{\sigma}} \leqslant \frac{L\delta_0}{2^t}
  \end{align}
  which completes the induction, our claim then follows immediately.
\end{proof}
\textbf{Gradient Complexity and LO Complexity:} This part is exactly the same as in convex case.

\subsection{Proof of Theorem \ref{thrm:STORC}}
\begin{proof}
Recall that in our setting even though $f(\cdot)$ is restricted strongly convex, we could have individual $f_i(\cdot)$ being non-convex. Our analysis starts with bounding the variance $\mathbb{E} \norm{\delta_k}^2$ differently from (\ref{vr_bound}) that requires convexity of $f_i(\cdot)$.

Suppose $\tilde{f}_i(x) = f_i(x) + \frac{\lambda}{2} \norm{x}^2$ is convex for some $\lambda$, note here by smoothness we can always choose $\lambda=L$, in our non-convex example, we can choose $\lambda =\lambda_{max}(\Sigma_w)$.
Notice that $\delta_k = \nabla f_i(z_k) - \nabla f_i(y_0) + \nabla f(y_0) -\nabla f(z_k)$, we define $\tilde{\delta_k} = \nabla \tilde{f}_i(z_k) -\nabla \tilde{f}_i(y_0) +\nabla \tilde{f}(y_0) -\nabla \tilde{f}(z_k)$. It is easy to
see that $\delta_k =\tilde{\delta}_k$, now we use convexity of $\tilde{f}(\cdot)$ to control the variance, notice that $\tilde{f}_i(\cdot)$ now is $L+\lambda$ smooth:
\begin{align}
  \mathbb{E} \norm{\delta_k}^2 &  =\mathbb{E} \norm{\tilde{\delta_k}}^2 \\
  & \leqslant 4(L+\lambda) \left[\tilde{f}(z_k) - f(\hat{\theta}) -\inner{\nabla \tilde{f} (\hat{\theta})}{z_k - \hat{\theta}}  + \tilde{f}(y_0) - f(\hat{\theta}) -\inner{\nabla \tilde{f} (\hat{\theta})}{y_0 - \hat{\theta}}\right] \\
\end{align}
We have
\begin{align}
  \tilde{f}(z_k) - f(\hat{\theta}) -\inner{\nabla \tilde{f} (\hat{\theta})}{z_k - \hat{\theta}} & = f(z_k) - f(\hat{\theta}) - \inner{\nabla f(\hat{\theta})}{z_k - \hat{\theta}} +\frac{\lambda}{2} \norm{z_k -\hat{\theta}}^2\\
  & \leqslant f(z_k) -f(\hat{\theta}) +\frac{\lambda}{2} \norm{z_k -\hat{\theta}}^2
\end{align}
by simple calculation, plug in this we have:
\begin{align}
    \mathbb{E} \norm{\delta_k}^2  \leqslant 4(L+\lambda) \left[f(z_k) - f(\hat{\theta}) + f(y_0) - f(\hat{\theta}) \right] + 2(L\lambda + \lambda^2) \left[\norm{z_k -\hat{\theta}}^2 +\norm{y_0 -\hat{\theta}}^2 \right]
\end{align}
Apply pseudo strong convexity (\ref{pseudo-sc}) again with assumption that we the algorithm has not converged to statistical precision we have:
\begin{align}\label{vr_nonconvex}
  \mathbb{E} \norm{\delta_k}^2 & \leqslant 4(L+\lambda) \left[f(z_k) - f(\hat{\theta}) + f(y_0) - f(\hat{\theta}) \right] + 4\frac{L\lambda + \lambda^2}{\hat{\sigma}} \left[f(z_k) - f(\hat{\theta}) + f(y_0) - f(\hat{\theta}) + 2\hat{\sigma} \epsilon_{stat}^2 \right] \\
  & \leqslant 4(L+\lambda)(1+\frac{\lambda}{\hat{\sigma}}) [f(z_k) -f(\hat{\theta})] + 12(L+\lambda)(1+\frac{\lambda}{\hat{\sigma}}) [ f(y_0) -f(\hat{\theta})] \\
  & = 4 \tilde{L} [f(z_k) -f(\hat{\theta})] + 12 \tilde{L} [ f(y_0) -f(\hat{\theta})]
\end{align}
where we have defined $\tilde{L} = (L+\lambda)(1+\frac{\lambda}{\hat{\sigma}})$.
We are now ready to complete our proof. Following the same argument of analysis of Non-convex case for determinstic CGS, we have modified Theorem \ref{cgs_convergence} as the following:
\begin{align}
  \mathbb{E}[f(y_k)-f(x)] & \leqslant  \Gamma_k \beta_1 \mathbb{E}\norm {x-x_0}^2  + \Gamma_k \sum_{i=1}^k \frac{\eta_{t,i} \gamma_i}{\Gamma_i} + \Gamma_k \sum_{i=1}^k \frac{\sigma_i^2\gamma_i}{\Gamma_i(\beta_i-L\gamma_i)} + \Gamma_k \sum_{i=1}^k \frac{\gamma_i \hat{\sigma} \epsilon_{stat}^2}{\Gamma_i}
\end{align}
follow the exact same argument as Corollary \ref{inner_convergence}, we obtain the modified version: if we can control $\sigma_i^2$ such that $\sigma_i^2 \leqslant \frac{L^2D_t^2}{N_t(i+1)^2}$ for all $i \leqslant k$, then we have:
\begin{align}\label{inner_convegence_nonconvex_stochastic}
  \mathbb{E} \left[f(y_k)-f(\hat{\theta}) \right]\leqslant \frac{8LD_t^2}{k(k+1)} + \hat{\sigma}\epsilon_{stat}^2 \leqslant \frac{16LD_t^2}{k(k+1)}
\end{align}
where the last inequality follows from the same argument as in (\ref{inner_convergence_nonconvex}).

We will follow the exact same induction argument in the proof of Theorem \ref{thrm:STORC} for convex case, with now a slight modification due to the difference of bounding the variance (\ref{vr_bound}) and (\ref{vr_nonconvex}).
For simplicity we will complete the induction step of showing $\mathbb{E} [f(y_t) -f(\hat{\theta})] \leqslant \frac{\delta_0}{2^t}$ provided $\mathbb{E} [f(y_{t-1}) -f(\hat{\theta})] \leqslant \frac{\delta_0}{2^{t-1}}$.
In $t$-th outer iteration, we will show by induction that (\ref{inner_convegence_nonconvex_stochastic}) holds for all $k$ and $\mathbb{E} [f(y_t) -f(\hat{\theta})] \leqslant \frac{\delta_0}{2^t}$ follows immediately for our choice of $N_t$.

Again from (\ref{precision_assumption}) which does not relies on convexity, we have:
\begin{align}
  \mathbb{E} [f(z_k) - f(\hat{\theta})] & \leqslant 10\frac{L}{\hat{\sigma}} \left (f(y_{k-1} -f(\hat{\theta})) \right) +8\frac{L}{\hat{\sigma}} \left (f(y_{k-2} -f(\hat{\theta})) \right) \\
  & \leqslant \frac{1200 \hat{u}LD_t^2}{(k+1)^2}
\end{align}
where the last inequality follows from inductive assumption on $f(y_{k-1})$ and $f(y_{k-2})$.

Keeping all the parameter same as before with $m_{t,k}$ to be decided,  applying variance bound (\ref{vr_nonconvex}) we obtain:
\begin{align}
  \sigma_k^2 &  = \frac{1}{m_{t,k}} \mathbb{E} \norm{\delta_k}^2  \\
  & \leqslant \frac{1}{m_{t,k}} \left( 4 \tilde{L} [f(z_k) -f(\hat{\theta})] + 12 \tilde{L} [ f(y_0) -f(\hat{\theta})] \right ) \\
  & \leqslant \frac{1}{m_{t,k}} \frac{8000 \hat{u} \tilde{L} L D_t^2}{(k+1)^2}
  \leqslant \frac{L^2D_t^2}{N_t(k+1)^2}
\end{align}
by choosing $m_{t,k} = \frac{8000 \tilde{L} \hat{u} N_t}{L}$, hence we have completed induction for (\ref{inner_convegence_nonconvex_stochastic}) and take $k=N_t$ and
with our choice of $N_t$ we obtain $\mathbb{E} [f(y_t) -f(\hat{\theta})] \leqslant \frac{\delta_0}{2^t}$, which completes the proof for outer iteration.

\textbf{Gradient Complexity and LO Complexity}
Since we only changed parameter $m_{t,k}$ the calls to the linear oracle is not affected, this bound is the same as in STORC for convex case. The calls of gradient evaluation is given by:
\begin{align}
  \sum_{t=1}^T (n+\sum_{k=1}^{N_t} m_{t,k} )& = \mathcal{O} \left( \sum_{t=1}^T (n+ \frac{\tilde{L}}{L} \hat{u} N_t^2 )\right ) \\
  & =\mathcal{O}  \left((n+ \frac{\tilde{L}L}{\hat{\sigma^2}} ) \log(\frac{\delta_0}{\epsilon}) \right)
\end{align}
\end{proof}

\end{document}